\documentclass{article}

\PassOptionsToPackage{numbers,compress}{natbib}

\usepackage[eandd, preprint]{author_kit_neurips/neurips_2026}

\usepackage[utf8]{inputenc}
\usepackage[T1]{fontenc}
\usepackage{hyperref}
\usepackage{url}
\usepackage{booktabs}
\usepackage{amsfonts}
\usepackage{nicefrac}
\usepackage{microtype}
\usepackage{xcolor}
\usepackage{graphicx}
\usepackage{multirow}
\usepackage{enumitem}
\usepackage{tabularx}
\usepackage{longtable}

\newcolumntype{Y}{>{\centering\arraybackslash}X}
\newcolumntype{P}[1]{>{\centering\arraybackslash}p{#1}}
\usepackage{placeins}
\usepackage{wrapfig}
\usepackage{tikz}
\usetikzlibrary{shapes.geometric, arrows.meta, positioning, fit, calc, backgrounds, decorations.pathreplacing}
\usepackage{pgfplots}
\pgfplotsset{compat=1.18}
\usepgfplotslibrary{groupplots}
\definecolor{colBurmese}{RGB}{214,39,40}
\definecolor{colLuxembourgish}{RGB}{255,127,14}
\definecolor{colAlbanian}{RGB}{44,160,44}
\definecolor{colSeychelles}{RGB}{31,119,180}
\definecolor{colArmenian}{RGB}{148,103,189}
\definecolor{colSwahili}{RGB}{140,86,75}
\definecolor{colGeorgian}{RGB}{227,119,194}
\definecolor{colBahraini}{RGB}{188,189,34}
\definecolor{colSamoan}{RGB}{23,190,207}
\definecolor{colLao}{RGB}{127,127,127}
\definecolor{colRomansh}{RGB}{218,165,32}
\usepackage{standalone}
\usepackage{tipa}

\usepackage{bm}
\usepackage[most]{tcolorbox}

\title{WorldSpeech: A Multilingual Speech Corpus\\from Around the World}

\author{
\begin{tabular}{c}
\vspace{0.5cm}
  Antonis Asonitis \quad Luca A. Lanzendörfer \quad Frédéric Berdoz \quad Roger Wattenhofer \\
  ETH Zurich \\
  \texttt{\{aasonitis, lanzendoerfer, fberdoz, wattenhofer\}@ethz.ch}
\end{tabular}
}

\begin{document}

\maketitle

\begin{abstract}
Automatic speech recognition (ASR) performs well for high-resource languages with abundant paired audio-transcript data, but its accuracy degrades sharply for most languages due to limited publicly available aligned data. To this end, we introduce \textsc{WorldSpeech}, a 24\,kHz multilingual speech corpus comprising 65k hours of aligned audio-transcript data across 76 languages, collected from diverse public sources including parliamentary proceedings, international broadcasts, and public-domain audiobooks. For 37 languages, \textsc{WorldSpeech} provides more than 200 hours of aligned speech, with 28 exceeding 500 hours and 24 surpassing 1k hours. Fine-tuning existing ASR models on \textsc{WorldSpeech} results in an average relative Word-Error-Rate reduction of 63.5\% across 11 typologically diverse languages.
\end{abstract}

\section{Introduction}

While multilingual Automatic Speech Recognition (ASR) has improved substantially on languages with sufficient public training data~\cite{radford2023whisper, pratap2024mms}, for the rest of the world's languages there is a significant gap between the current publicly available aligned data and the volume of data needed to train ASR models to achieve high transcription accuracy. To close this gap, more paired speech data is required, and collecting this data for low-resource languages is inherently difficult.

Existing works have made progress to address the data scarcity of low-resource languages. Common Voice~\cite{ardila2020common} is the largest crowd-sourced public corpus of read speech with clean transcripts across many languages, but per-language hours remain limited. 
Audiobook collections~\cite{pratap2020mls} contain a larger number of hours but cover only a limited number of languages. Other approaches, such as MOSEL~\cite{gaido2024mosel} and YODAS~\cite{li2023yodas} reach hundreds of thousands of hours of audio, but their transcripts are pseudo-labels generated by ASR rather than human ground truth. A promising direction has been to gather data from institutions that systematically publish human-transcribed speech. VoxPopuli~\cite{wang2021voxpopuli}, ParlaSpeech~\cite{parlaspeech2022}, and EuroSpeech~\cite{pfisterer2025eurospeech} assemble tens of thousands of hours by aligning parliamentary recordings with verbatim transcripts. We find there are two aspects that restrict this approach from generalizing beyond Europe. First, many governments restrict access, publish no transcripts, or take recordings offline. Second, alignment yield collapses when the initial ASR transcribes the target language poorly, and therefore the alignment approach cannot find the ground-truth transcript belonging to the speech utterance. In this work we tackle both by finding new data sources for various languages and iteratively align audio recordings with transcripts, improving the number of available hours of paired data for low-resource languages.

\begin{figure}[!p]
\centering
\includegraphics[width=\textwidth]{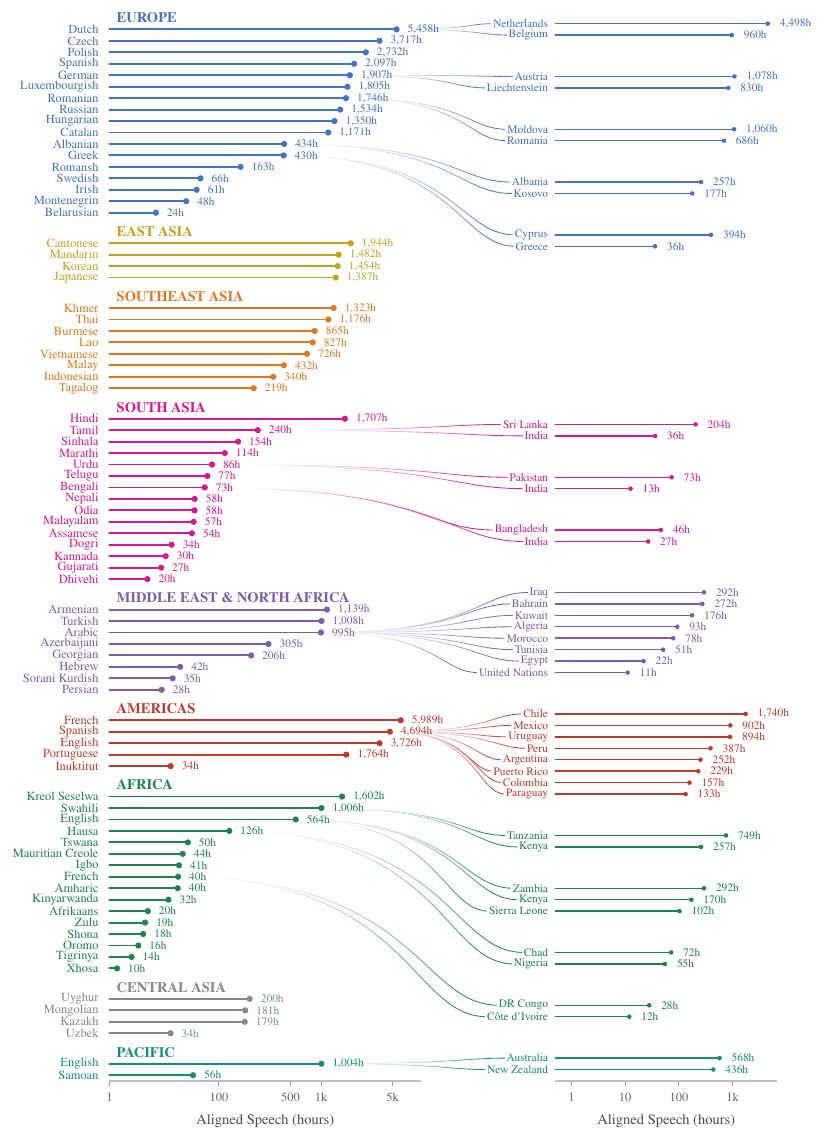}
\caption{Aligned-speech distribution across the languages in \textsc{WorldSpeech}. The left panel shows one row per language, with bar length giving total aligned hours on a log axis. Languages are grouped by continent (colored headers). The right panel decomposes languages with multiple country sources into per-country language variants.}
\label{fig:dataset_overview}
\end{figure}

We introduce \textsc{WorldSpeech}, a multilingual speech corpus, that addresses both data scarcity and poor ASR performance for various low-resource languages and language variants (see Figure~\ref{fig:dataset_overview} for overview). We extend the parliamentary recipe to international broadcasters with low-resource mandates, national public broadcasters, and public-domain audiobook archives. \textsc{WorldSpeech} contains 64{,}970 aligned hours across 76 languages. We use an alignment strategy~\cite{pfisterer2025eurospeech} to obtain matching utterance-transcript pairs from raw audio recordings and full session transcripts. For languages where ASR quality is the bottleneck rather than full transcript recording availability, we show that fine-tuning the initial ASR on the first-pass yield and re-aligning the same audio with this fine-tuned ASR recovers segments that initially were not matched, increasing the corpus by $+19.5\%$ to $+201.1\%$ per language without any new data collection. Fine-tuning a multilingual ASR backbone on these aligned hours reduces the Word-Error-Rate (WER) by $40.2\%$ to $91.7\%$ across 11 typologically diverse languages, with an average relative reduction of $63.5\%$.

\smallskip \noindent Our contributions can be summarized as follows:
\begin{itemize}
\item We introduce \textsc{WorldSpeech}, a multilingual speech corpus containing 65k aligned hours across 76 languages, with 24 languages above 1k hours, 28 above 500 hours, and 37 above 200 hours.\footnote{\url{https://huggingface.co/datasets/disco-eth/WorldSpeech}.}
\item We improve ASR performance by fine-tuning on \textsc{WorldSpeech}, reducing WER by $40\%$ to $92\%$ across 11 typologically diverse languages, with an average relative reduction of $63.5\%$.
\item For languages with weak initial ASR performance, we use an iterative alignment refinement scheme that fine-tunes the ASR model on segments matched in a first pass and realigns the corpus, yielding between $19.5\%$ and $201.1\%$ additional aligned data without requiring more data gathering.
\end{itemize}

\section{Related work}

\paragraph{Multilingual aligned-speech datasets.}
There exist many public and private multilingual datasets of paired audio-transcript data for ASR training. In the following we list and consolidate previous work (see Table~\ref{tab:dataset_comparison}). Public read-speech corpora, especially Common Voice~\cite{ardila2020common} and MLS~\cite{pratap2020mls} cover many languages, but provide limited per-language hours outside of a handful of well-resourced cases. Spontaneous-speech corpora have grown in popularity and use, examples include Emilia~\cite{he2024emilia}, MSR-86k~\cite{li2024msr86k}, GigaSpeech~\cite{chen2021gigaspeech} and GigaSpeech~2~\cite{yang2024gigaspeech2}, ReazonSpeech~\cite{yin2023reazonspeech}, and WenetSpeech~\cite{zhang2022wenetspeech}). Datasets built on parliamentary sources such as VoxPopuli~\cite{wang2021voxpopuli}, ParlaSpeech~\cite{parlaspeech2022}, and EuroSpeech~\cite{pfisterer2025eurospeech} provide human-annotated ground truth transcripts. The parliamentary datasets provides verbatim human transcripts and currently lead in per-language depth, but coverage has mainly focused on the European Union. Recent multilingual efforts include Granary~\cite{koluguri2025granary} containing 25 European languages with pseudo-labeled transcripts, NaijaVoices~\cite{naijavoices2025} with 3 Nigerian languages, of read-speech, Speech-MASSIVE~\cite{lee2024speechmassive} containing 12 languages, and the OWSM training assemblage~\cite{peng2023owsm} with 151 languages drawn from existing labeled corpora. YODAS~\cite{li2023yodas} reaches 149 languages but contains generated rather than human-annotated transcripts, VoxLingua107~\cite{valk2021voxlingua107} contains 107 languages of YouTube speech labelled by spoken language for spoken-language identification (no transcripts), and FLEURS~\cite{conneau2022fleurs} has 102 languages with an average of 12h per language. Single-language heavy datasets such as Libriheavy~\cite{khan2024libriheavy} and AfriSpeech-200~\cite{olatunji2023afrispeech} cover one language deeply rather than many. The largest spontaneous corpora overall are not publicly accessible (such as Whisper Data~\cite{radford2023whisper}, MMS-Lab~\cite{pratap2024mms}, SeamlessM4T~\cite{barrault2023seamless}, BABEL~\cite{babel2014}). \textsc{WorldSpeech} is, to our knowledge, the first publicly available corpus to combine per-language depth, with 24 languages exceeding 1k hours of human-labeled data, with broad coverage across 76 languages and their regional varieties.

\paragraph{Aligning audio to in-the-wild transcripts.}
Classical forced aligners~\cite{mfa2017} require pronunciation dictionaries and acoustic models that are not available for the majority of our target languages: MFA releases pretrained models for 41 languages, none of which include Kreol Seselwa, Romansh, Luxembourgish, Khmer, Lao, Sinhala, Burmese, Tagalog, Cantonese and other low-resource languages we target. ASR-based pipelines~\cite{bain2023whisperx,pfisterer2025eurospeech,kuerzinger2020ctcseg} sidestep this by matching ASR output against the human transcript via edit distance, dynamic time warping, or CTC-based segmentation, requiring only an ASR model rather than a pronunciation lexicon. We build on EuroSpeech's two-stage coarse-to-fine variant for our long-audio long-transcript strategy (see Table~\ref{sec:pipeline}).

\paragraph{Iterative self-training and pseudo-labeling.}
Iterative pseudo-labeling (IPL)~\cite{xu2020iterative} and its multilingual extension~\cite{lugosch2022pseudo} alternate between training an ASR on its current labeled set and using that ASR to generate transcripts for additional unlabeled audio, expanding the training set with model-generated labels at each iteration. Whisper~\cite{radford2023whisper} applies the same principle at scale, using its v2 model to pseudo-label audio for v3 training. Our iterative alignment refinement (see Table~\ref{sec:iterative}) shares the iterative loop structure but operates in a different setting, where human transcripts are already paired with the audio and the initial ASR is iteratively fine-tuned to match the audio more reliably to the transcripts through our alignment pipeline.

\begin{table}[!t]
\centering
\small
\setlength{\tabcolsep}{4pt}
\renewcommand{\arraystretch}{0.95}
\caption{Comparison of multilingual aligned-speech datasets. We define thresholds on number of languages with at least 50, 200, 500, or 1k hours of paired audio. ``-'' marks values not reported in the cited source. Datasets are grouped by redistribution status and transcript quality, and within each group rows are sorted by total hours descending.}
\label{tab:dataset_comparison}
\begin{tabularx}{\textwidth}{l Y Y Y Y Y Y}
\toprule
Dataset                                       & Total hours & \# Lang. & $\geq$50\,h & $\geq$200\,h & $\geq$500\,h & $\geq$1{,}000\,h \\
\midrule
\multicolumn{7}{@{}l}{\textit{Private or restricted}} \\
\midrule
SeamlessM4T~\cite{barrault2023seamless}       & $443.0$k    & $35$        & -             & -              & -              & -              \\
Whisper Data~\cite{radford2023whisper}        & $117.0$k    & $96$        & $35$            & $25$             & $21$             & -              \\
MMS-Lab~\cite{pratap2024mms}                  & $44.7$k     & $1107$      & -             & $0$              & $0$              & $0$              \\
\midrule
\multicolumn{7}{@{}l}{\textit{Public, auto-generated transcripts}} \\
\midrule
MOSEL (unlabeled)~\cite{gaido2024mosel}      & $950.0$k    & $24$        & $23$            & $23$             & $23$             & $23$             \\
YODAS~\cite{li2023yodas}                      & $369.5$k    & $149$       & $24$            & $18$             & $15$             & $13$             \\
Emilia~\cite{he2024emilia}                    & $101.0$k    & $6$         & $6$             & $6$              & $5$              & $5$              \\
MSR-86k~\cite{li2024msr86k}                   & $86.3$k     & $15$        & $15$            & $15$             & $15$             & $14$             \\
GigaSpeech 2~\cite{yang2024gigaspeech2}       & $30.0$k     & $3$         & $3$             & $3$              & $3$              & $3$              \\
\midrule
\multicolumn{7}{@{}l}{\textit{Public, ground-truth aligned}} \\
\midrule
EuroSpeech~\cite{pfisterer2025eurospeech}     & $61.0$k     & $22$        & $22$            & $22$             & $22$             & $20$          \\
MLS~\cite{pratap2020mls}                      & $50.0$k     & $8$         & $8$             & $6$              & $5$              & $4$              \\
Common Voice~\cite{ardila2020common}          & $22.1$k     & $134$       & $45$            & $23$             & $15$             & $8$              \\
CMU Wilderness~\cite{black2019wilderness}     & $14.0$k     & \textbf{700}       & $30$            & $0$              & $0$              & $0$              \\
VoxPopuli~\cite{wang2021voxpopuli}            & $1.8$k      & $16$        & $10$            & $3$              & $1$              & $0$              \\
FLEURS~\cite{conneau2022fleurs}               & $1.4$k      & $102$       & $0$             & $0$              & $0$              & $0$              \\
\midrule
\textbf{\textsc{WorldSpeech}}                 & $\mathbf{65}$\textbf{k} & 76 & \textbf{53} & \textbf{37} & \textbf{28} & \textbf{24} \\
\bottomrule
 \end{tabularx}
\end{table}

\section{Alignment pipeline}
\label{sec:pipeline}

\paragraph{Data collection and standardization.}
Each source releases audio and transcripts in its own combination of formats, and the per-source preprocessing dominates the engineering effort. Audio is downloaded as MP4 or MP3 files, retrieved from YouTube, captured from HLS streams at session granularity, or concatenated from multi-part files. Transcripts are sourced from HTML and XML APIs, DOCX files, SRT subtitles, and PDFs. We standardize all audio to mono 24\,kHz and all transcripts to plain text. PDF transcripts can introduce several failure modes that require different solutions depending on country and formatting style. Two-column layouts, common in African and Asian Hansard-style documents, often cause standard text extraction to interleave content across both columns. We detect these layouts by rendering pages as images, then crop and extract each column independently, processing the right column first for right-to-left scripts such as Arabic. Some PDFs use font encodings that render correctly on screen but produce wrong Unicode under programmatic extraction, in which case we fall back to OCR. Tesseract~\cite{smith2007tesseract} is used for Latin, Cyrillic, Greek, and Arabic scripts, and the Surya neural OCR engine~\cite{surya2024} for scripts where Tesseract performs poorly (e.g., Nastaliq Urdu and Burmese, where cursive ligatures and consonant stacking cause systematic character-segmentation errors). Further language-specific normalization is applied, including stripping Arabic harakat, normalizing alef and teh-marbuta forms, removing Cyrillic page headers, romanization and tone-mark normalization for Cantonese transcripts, and handling intra-session code-switching in multilingual parliaments such as the Philippines, South Africa and Pakistan, by first running language detection using Whisper-large-v3-turbo and then transcribing using the corresponding language token. The collection and preprocessing pipelines cover 79 distinct parliamentary and public-domain sources across 82 countries. 

\paragraph{Segmentation and ASR.}
After transcript extraction, we segment long-form audio (1-10 hours per session) into short utterances suitable for alignment. We use Silero VAD~\cite{silero2024} to detect speech regions, then apply a sliding-window segmentation that cuts at natural silences to produce segments of 3-30 seconds. Each segment is transcribed by an ASR model selected through an ablation, where up to three candidate models are run on 10 hours of the target language's audio and the one that maximizes the fraction of segments passing the CER$<$0.3 threshold is chosen for the full run. In practice, Whisper-large-v3-turbo~\cite{radford2023whisper} suffices for most European and widely-resourced languages, while MMS-1B~\cite{pratap2024mms} with per-language adapters is better-suited for languages where Whisper produces script errors or hallucinated output, and a community fine-tune is used where one exists. The chosen model per language is reported in Appendix~\ref{app:asr_alignment_models} (see Table~\ref{tab:asr_alignment_models}).

\paragraph{Audio-transcript pairing.}
Before we can run audio-transcript alignment, each audio file must be paired with its corresponding transcript. Due to the inconsistencies of data formats among the parsed sources, this step is non-trivial. Parliamentary sources that publish verbatim records do so through a variety of mechanisms, some expose a structured API where session identifiers appear in both the video metadata and the transcript URL, others require gathering information from a calendar or agenda page to extract the session date, then cross-referencing it against a separately maintained transcript archive which can be hosted on a completely different domain. Dates are frequently missing, inconsistent between audio and transcript systems, or present in different formats across years of the same parliament. For some sources the only reliable key is the session title or a sequential document number, requiring fuzzy string matching. Broadcaster archives such as Radio Free Asia (RFA) and Voice of America (VOA) present a different problem. Each article page carries both the audio file and the article text, so pairing is automatic, but article publication timestamps do not always match the broadcast date. In all cases, pairing is validated by running ASR on a short sample of the audio and verifying that the output shares vocabulary with the transcript before committing to a full download.

\paragraph{CER-based matching.}
We reuse the two-stage coarse-to-fine CER alignment of EuroSpeech~\cite{pfisterer2025eurospeech} as the inner matching loop. Each VAD segment is transcribed by the ASR model to produce a character hypothesis. A sliding window advances over the human transcript, computing the character error rate (CER) between the hypothesis and each candidate span. The span minimizing CER is selected and the segment is retained with the human-transcript text as its label if this minimum CER falls below 0.3. The ASR output is used only for search and is never stored as ground truth. The search space varies by source type. For parliamentary sessions the window searches the full verbatim document using the two-stage coarse-to-fine strategy introduced in \cite{pfisterer2025eurospeech}. For broadcaster archives where each clip arrives with its own transcript, the search is restricted to that transcript. Where only partial transcripts exist, such as agenda-only committee minutes or news bulletins that do not transcribe the full broadcast, the window searches each available fragment and audio regions between fragments are left unaligned. For audiobooks, the long-audio strategy is applied per chapter after chapter-boundary metadata constrains the search. The per-segment CER is stored as metadata so users can filter the dataset according to their own threshold.

Matching yield is the most failure-prone part of the pipeline, degrading sharply when the intitial ASR is weak on the target language. A hypothesis transcription full of hallucinated or script-wrong characters will not align with any window below the threshold even when the correct passage exists in the transcript. For languages such as Burmese, Khmer, Lao, and Sinhala, initial yield with the baseline ASR models covered only a fraction of the available audio, even after verified pairing confirmed that the transcripts were present and correctly formatted. This is the core failure mode that motivates the iterative refinement described in Section~\ref{sec:iterative}.

\section{The \textsc{WorldSpeech} Dataset}
\label{sec:dataset}

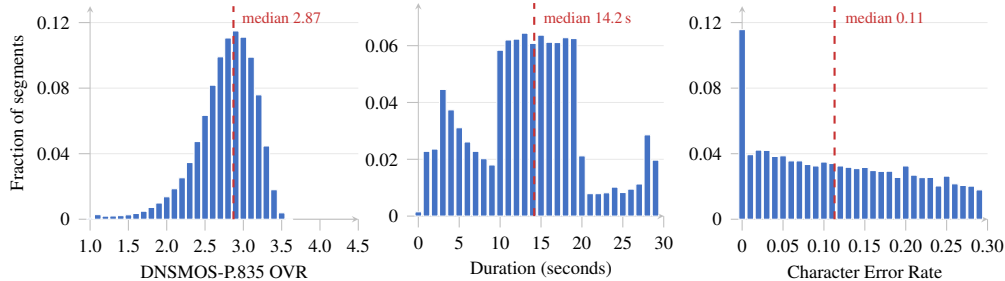
\begin{figure}[!t]
\centering
\definecolor{histBlue}{HTML}{4472C4}
\definecolor{histRed}{RGB}{200,50,50}
\pgfplotsset{
  histStyle/.style={
    ybar,
    fill=histBlue, draw=histBlue!85!black, line width=0.1pt,
    width=1.08\linewidth, height=4.4cm,
    axis x line=bottom,
    axis y line=left,
    x axis line style={color=gray!50},
    y axis line style={color=gray!50},
    tick style={color=gray!50, thin},
    tick align=outside,
    tick label style={font=\scriptsize, /pgf/number format/assume math mode=true},
    xticklabel style={anchor=north},
    xlabel style={font=\scriptsize, yshift=2pt},
    ylabel style={font=\scriptsize, yshift=-3pt},
    ymajorgrids=true,
    xmajorgrids=false,
    grid style={gray!18, very thin},
    enlarge x limits=false,
    scaled y ticks=false,
    bar shift=0pt,
    clip=false,
  }
}
\noindent
\begin{minipage}[b]{0.34\textwidth}
\centering
\begin{tikzpicture}
\begin{axis}[
  histStyle,
  xmin=1.0, xmax=4.5,
  ymin=0, ymax=0.13,
  xlabel={DNSMOS-P.835 OVR},
  ylabel={Fraction of segments},
  xtick={1.0,1.5,2.0,2.5,3.0,3.5,4.0,4.5},
  xticklabels={1.0,1.5,2.0,2.5,3.0,3.5,4.0,4.5},
  xticklabel style={anchor=north},
  enlarge x limits=false,
  ytick={0,0.04,0.08,0.12},
  yticklabels={0,0.04,0.08,0.12},
]
\addplot[fill=histBlue, draw=white, line width=0.4pt, fill opacity=1, bar width=2.5pt] coordinates {
  (1.0,0.0007)(1.1,0.0028)(1.2,0.0019)(1.3,0.0020)(1.4,0.0022)
  (1.5,0.0027)(1.6,0.0035)(1.7,0.0049)(1.8,0.0072)(1.9,0.0099)
  (2.0,0.0137)(2.1,0.0186)(2.2,0.0253)(2.3,0.0346)(2.4,0.0474)
  (2.5,0.0634)(2.6,0.0818)(2.7,0.0991)(2.8,0.1106)(2.9,0.1149)
  (3.0,0.1111)(3.1,0.0989)(3.2,0.0759)(3.3,0.0447)(3.4,0.0180)
  (3.5,0.0039)(3.6,0.0002)
};
\draw[histRed, dashed, line width=0.8pt] (axis cs:2.871,0) -- (axis cs:2.871,0.13);
\node[font=\tiny, color=histRed, anchor=south west, inner sep=1pt]
  at (axis cs:2.95,0.118) {median 2.87};
\end{axis}
\end{tikzpicture}
\end{minipage}\hspace{-4pt}%
\begin{minipage}[b]{0.32\textwidth}
\centering
\begin{tikzpicture}
\begin{axis}[
  histStyle,
  xmin=0, xmax=30,
  ymin=0, ymax=0.075,
  xlabel={Duration (seconds)},
  xtick={0,5,10,15,20,25,30},
  xticklabels={0,5,10,15,20,25,30},
  xticklabel style={anchor=north},
  enlarge x limits=false,
  ytick={0,0.02,0.04,0.06},
  yticklabels={0,0.02,0.04,0.06},
]
\addplot[fill=histBlue, draw=white, line width=0.4pt, fill opacity=1, bar width=2.5pt] coordinates {
  (0,0.0015)(1,0.0228)(2,0.0236)(3,0.0446)(4,0.0374)
  (5,0.0311)(6,0.0261)(7,0.0228)(8,0.0202)(9,0.0180)
  (10,0.0584)(11,0.0620)(12,0.0623)(13,0.0644)(14,0.0608)
  (15,0.0637)(16,0.0612)(17,0.0611)(18,0.0628)(19,0.0625)
  (20,0.0212)(21,0.0079)(22,0.0079)(23,0.0082)(24,0.0102)
  (25,0.0083)(26,0.0095)(27,0.0113)(28,0.0286)(29,0.0197)
};
\draw[histRed, dashed, line width=0.8pt] (axis cs:14.17,0) -- (axis cs:14.17,0.075);
\node[font=\tiny, color=histRed, anchor=south west, inner sep=1pt]
  at (axis cs:15.0,0.067) {median 14.2\,s};
\end{axis}
\end{tikzpicture}
\end{minipage}\hspace{-4pt}%
\begin{minipage}[b]{0.32\textwidth}
\centering
\begin{tikzpicture}
\begin{axis}[
  histStyle,
  xmin=0, xmax=0.30,
  ymin=0, ymax=0.13,
  xlabel={Character Error Rate},
  xtick={0,0.05,0.10,0.15,0.20,0.25,0.30},
  xticklabels={0,0.05,0.10,0.15,0.20,0.25,0.30},
  xticklabel style={anchor=north},
  enlarge x limits=false,
  ytick={0,0.04,0.08,0.12},
  yticklabels={0,0.04,0.08,0.12},
]
\addplot[fill=histBlue, draw=white, line width=0.4pt, fill opacity=1, bar width=2.5pt] coordinates {
  (0.00,0.1157)(0.01,0.0394)(0.02,0.0422)(0.03,0.0420)(0.04,0.0383)
  (0.05,0.0387)(0.06,0.0356)(0.07,0.0356)(0.08,0.0334)(0.09,0.0325)
  (0.10,0.0349)(0.11,0.0340)(0.12,0.0325)(0.13,0.0317)(0.14,0.0309)
  (0.15,0.0316)(0.16,0.0297)(0.17,0.0292)(0.18,0.0293)(0.19,0.0255)
  (0.20,0.0325)(0.21,0.0269)(0.22,0.0254)(0.23,0.0255)(0.24,0.0203)
  (0.25,0.0263)(0.26,0.0216)(0.27,0.0206)(0.28,0.0202)(0.29,0.0179)
};
\draw[histRed, dashed, line width=0.8pt] (axis cs:0.113,0) -- (axis cs:0.113,0.13);
\node[font=\tiny, color=histRed, anchor=south west, inner sep=1pt]
  at (axis cs:0.123,0.118) {median 0.11};
\end{axis}
\end{tikzpicture}
\end{minipage}

 \caption{Corpus-wide unit-normalized distributions across the aligned segments of \textsc{WorldSpeech}. \emph{Left:} audio quality. \emph{Centre:} segment duration in seconds. Our alignment pipeline merges or cuts segments into roughly 10-20\,s duration. The bump at 3-4\,s comes from short standalone utterances obtained after VAD. The pile-up at 28-29\,s comes from the 30\,s cutoff threshold forcing a cut when no silence is found inside the segmentation window. \emph{Right:} alignment character error rate, median 0.11, the 0.30 endpoint is the cutoff threshold.} 
\label{fig:quality_histograms}
\end{figure}

Applying our alignment pipeline to parliamentary recordings and public-domain sources, we construct \textsc{WorldSpeech}, a multilingual aligned speech dataset of 64{,}970 hours covering 76 languages (Table~\ref{tab:dataset_comparison}). The configurations span from low-resource languages for which no prior open-source aligned corpus existed to dialect variants of widely spoken languages that lack dialect-specific training corpora like Qu\'eb\'ecois French, Austrian and Liechtenstein German, six Latin American Spanish varieties, Bahraini and Moroccan Arabic.

\paragraph{Composition and coverage.}
Table~\ref{tab:dataset_comparison} compares \textsc{WorldSpeech} against existing multilingual aligned-speech datasets, and Figure~\ref{fig:dataset_overview} gives the per-language breakdown grouped by region for the 76 languages with at least 10 hours of aligned data. The majority of data comes from national parliamentary archives, complemented by national public broadcasters, international public-service broadcasters (RFA, VOA, RFE/RL, all broadcast news and interviews),\footnote{Voice of America (VOA), Radio Free Asia (RFA), Radio Free Europe (RFE) and Radio Liberty are United States federal grantee broadcasters whose statutory mandate is to provide independent news in diverse languages to audiences in restricted-media environments, including many languages otherwise under-represented in commercial broadcasting (e.g., Khmer, Lao, Burmese, Uyghur, Tigrinya)} public-domain audiobooks (LibriVox~\cite{librivox}, Aozora~\cite{aozora}, Ben-Yehuda~\cite{benyehuda}, all read literature), and a small set of other public-domain sources (see full source list in Appendix~\ref{app:sources}). Where multiple source types are available for the same language, this mix spans up to three register types (formal prepared, broadcast journalistic, read literary), which reduces the single-source bias inherent in parliament-only corpora. To the best of our knowledge, for 48 of the included languages, \textsc{WorldSpeech} constitutes the largest or only publicly available ground-truth aligned corpus. The full comparison to previous work with hours per language can be found in Appendix~\ref{app:prior_census}, and details on the source licenses can be found in Appendix~\ref{app:licenses}. An additional feature beyond scale is the coverage of regional and dialectal variants that are often absent from existing corpora, supporting dialect-specific evaluation and fine-tuning.

\paragraph{Format, quality tiers, and metadata.}

Each row in the dataset contains audio segments up to 30 seconds in duration combined with the ground-truth transcript, the ASR transcript used during alignment, the segment-level CER between the two, the language code, the duration, the source identifier, and the session date. Each row also contains a DNSMOS~\cite{reddy2022dnsmos} audio-quality score and a signal-to-noise-ratio estimate, which lets users filter the corpus to a high-quality subset without further preprocessing. 
The dataset is organised into one configuration per country-language pair, with a 95/5 train/test split.

Figure~\ref{fig:quality_histograms} reports the corpus-wide distributions of audio quality, segment duration, and alignment CER over the retained subset. DNSMOS-P.835 OVR has median 2.87, consistent with the broadcast and parliamentary origin of most sources. The segment duration distribution is shaped by the alignment pipeline's 10-20\,s target window, with median 14.2\,s. The small bump at 3-4\,s stems from short standalone utterances isolated by VAD, and the pile-up at 28-29\,s stems from the 30\,s upper bound forcing a cut when no silence is found inside the window. CER has median 0.11, with a peak at 0.00 (around 11.6\% of retained segments, where the alignment ASR transcribes the audio identically to the human transcript) and a long flat tail between 0.01 and 0.30 reflecting usable but imperfect alignments. Users can modify the filter at any threshold using the per-segment CER value provided in the dataset (Table~\ref{tab:dataset_comparison}).

\paragraph{Filtering the corpus into task-specific subsets.}
The metadata associated with each segment lets users construct task-specific subsets without re-processing the audio. Users can filter on CER to obtain ASR training subsets at different quality levels. Filtering on session metadata isolates potential sources that contain code-switching from multilingual jurisdictions (the Philippines House and Senate, the South African Parliament, the Belgian chambers, and the Pakistani sessions) and dialect-specific subsets (Latin American Spanish across seven countries, Arabic across seven jurisdictions, Hindi across four Indian states, English across six countries) for dialect-aware training and evaluation. 

\section{Experiments}
\label{sec:experiments}

To evaluate the quality of \textsc{WorldSpeech}, we fine-tune an ASR model on our data for typologically diverse languages and measure word error rate (WER) and character error rate (CER) on public benchmarks where available and on the \textsc{WorldSpeech} held-out test sets when no available test benchmark exists.

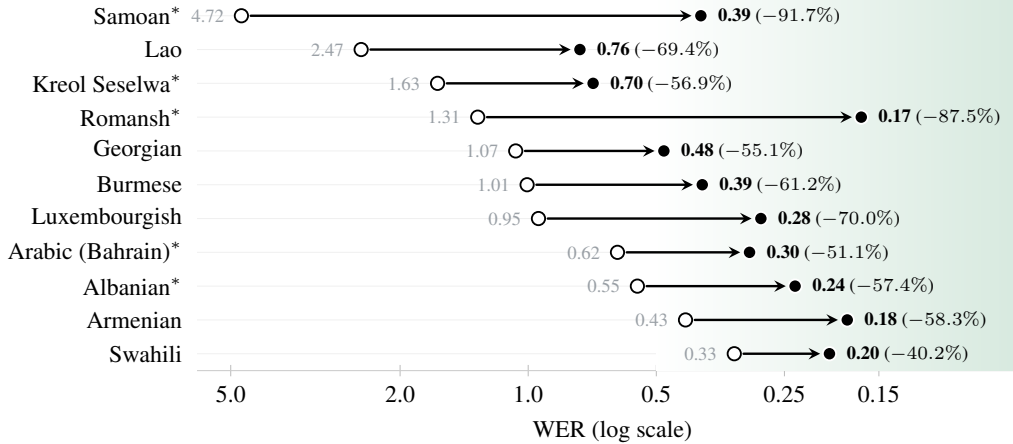
\begin{figure}[t]
\centering
\definecolor{asrGray}{RGB}{154,160,166}
\definecolor{asrGreen}{HTML}{1E8449}
\begin{tikzpicture}[>=stealth]
\begin{axis}[
  xmode=log,
  x dir=reverse,
  xmin=0.067, xmax=6.0,
  ymin=-1.5, ymax=9.5,
  ytick={-1,0,1,2,3,4,5,6,7,8,9},
  yticklabels={
    {Swahili},
    {Armenian},
    {Albanian$^{*}$},
    {Arabic (Bahrain)$^{*}$},
    {Luxembourgish},
    {Burmese},
    {Georgian},
    {Romansh$^{*}$},
    {Kreol Seselwa$^{*}$},
    {Lao},
    {Samoan$^{*}$}
  },
  tick label style={font=\small},
  ytick style={draw=none},
  xtick={0.15,0.25,0.5,1.0,2.0,5.0},
  xticklabels={0.15,0.25,0.5,1.0,2.0,5.0},
  xtick style={color=gray!50,thin},
  xlabel={WER (log scale)},
  xlabel style={font=\small},
  axis x line=bottom,
  axis y line=left,
  x axis line style={color=gray!40, -},
  y axis line style={draw=none},
  ymajorgrids=true,
  grid style={gray!12,very thin},
  width=0.9\textwidth,
  height=6.5cm,
  clip=false,
]
\shade[left color=white, right color=asrGreen!18!white]
  (axis cs:0.5,-1.5) rectangle (axis cs:0.067,9.5);
\draw[black,line width=0.9pt,line cap=round,->,shorten >=3.5pt,shorten <=3.5pt]
  (axis cs:4.719,9)--(axis cs:0.393,9);
\filldraw[fill=white,draw=black,line width=0.7pt](axis cs:4.719,9)circle(2.5pt);
\filldraw[fill=black,draw=white,line width=0.7pt](axis cs:0.393,9)circle(2.5pt);
\node[anchor=east,xshift=-3pt,font=\scriptsize,color=asrGray]at(axis cs:4.719,9){4.72};
\node[anchor=west,xshift=3pt,font=\scriptsize]at(axis cs:0.393,9){\textbf{0.39} ($-91.7\%$)};
\draw[black,line width=0.9pt,line cap=round,->,shorten >=3.5pt,shorten <=3.5pt]
  (axis cs:2.469,8)--(axis cs:0.756,8);
\filldraw[fill=white,draw=black,line width=0.7pt](axis cs:2.469,8)circle(2.5pt);
\filldraw[fill=black,draw=white,line width=0.7pt](axis cs:0.756,8)circle(2.5pt);
\node[anchor=east,xshift=-3pt,font=\scriptsize,color=asrGray]at(axis cs:2.469,8){2.47};
\node[anchor=west,xshift=3pt,font=\scriptsize]at(axis cs:0.756,8){\textbf{0.76} ($-69.4\%$)};
\draw[black,line width=0.9pt,line cap=round,->,shorten >=3.5pt,shorten <=3.5pt]
  (axis cs:1.633,7)--(axis cs:0.704,7);
\filldraw[fill=white,draw=black,line width=0.7pt](axis cs:1.633,7)circle(2.5pt);
\filldraw[fill=black,draw=white,line width=0.7pt](axis cs:0.704,7)circle(2.5pt);
\node[anchor=east,xshift=-3pt,font=\scriptsize,color=asrGray]at(axis cs:1.633,7){1.63};
\node[anchor=west,xshift=3pt,font=\scriptsize]at(axis cs:0.704,7){\textbf{0.70} ($-56.9\%$)};
\draw[black,line width=0.9pt,line cap=round,->,shorten >=3.5pt,shorten <=3.5pt]
  (axis cs:1.314,6)--(axis cs:0.165,6);
\filldraw[fill=white,draw=black,line width=0.7pt](axis cs:1.314,6)circle(2.5pt);
\filldraw[fill=black,draw=white,line width=0.7pt](axis cs:0.165,6)circle(2.5pt);
\node[anchor=east,xshift=-3pt,font=\scriptsize,color=asrGray]at(axis cs:1.314,6){1.31};
\node[anchor=west,xshift=3pt,font=\scriptsize]at(axis cs:0.165,6){\textbf{0.17} ($-87.5\%$)};
\draw[black,line width=0.9pt,line cap=round,->,shorten >=3.5pt,shorten <=3.5pt]
  (axis cs:1.070,5)--(axis cs:0.480,5);
\filldraw[fill=white,draw=black,line width=0.7pt](axis cs:1.070,5)circle(2.5pt);
\filldraw[fill=black,draw=white,line width=0.7pt](axis cs:0.480,5)circle(2.5pt);
\node[anchor=east,xshift=-3pt,font=\scriptsize,color=asrGray]at(axis cs:1.070,5){1.07};
\node[anchor=west,xshift=3pt,font=\scriptsize]at(axis cs:0.480,5){\textbf{0.48} ($-55.1\%$)};
\draw[black,line width=0.9pt,line cap=round,->,shorten >=3.5pt,shorten <=3.5pt]
  (axis cs:1.006,4)--(axis cs:0.390,4);
\filldraw[fill=white,draw=black,line width=0.7pt](axis cs:1.006,4)circle(2.5pt);
\filldraw[fill=black,draw=white,line width=0.7pt](axis cs:0.390,4)circle(2.5pt);
\node[anchor=east,xshift=-3pt,font=\scriptsize,color=asrGray]at(axis cs:1.006,4){1.01};
\node[anchor=west,xshift=3pt,font=\scriptsize]at(axis cs:0.390,4){\textbf{0.39} ($-61.2\%$)};
\draw[black,line width=0.9pt,line cap=round,->,shorten >=3.5pt,shorten <=3.5pt]
  (axis cs:0.946,3)--(axis cs:0.284,3);
\filldraw[fill=white,draw=black,line width=0.7pt](axis cs:0.946,3)circle(2.5pt);
\filldraw[fill=black,draw=white,line width=0.7pt](axis cs:0.284,3)circle(2.5pt);
\node[anchor=east,xshift=-3pt,font=\scriptsize,color=asrGray]at(axis cs:0.946,3){0.95};
\node[anchor=west,xshift=3pt,font=\scriptsize]at(axis cs:0.284,3){\textbf{0.28} ($-70.0\%$)};
\draw[black,line width=0.9pt,line cap=round,->,shorten >=3.5pt,shorten <=3.5pt]
  (axis cs:0.617,2)--(axis cs:0.302,2);
\filldraw[fill=white,draw=black,line width=0.7pt](axis cs:0.617,2)circle(2.5pt);
\filldraw[fill=black,draw=white,line width=0.7pt](axis cs:0.302,2)circle(2.5pt);
\node[anchor=east,xshift=-3pt,font=\scriptsize,color=asrGray]at(axis cs:0.617,2){0.62};
\node[anchor=west,xshift=3pt,font=\scriptsize]at(axis cs:0.302,2){\textbf{0.30} ($-51.1\%$)};
\draw[black,line width=0.9pt,line cap=round,->,shorten >=3.5pt,shorten <=3.5pt]
  (axis cs:0.554,1)--(axis cs:0.236,1);
\filldraw[fill=white,draw=black,line width=0.7pt](axis cs:0.554,1)circle(2.5pt);
\filldraw[fill=black,draw=white,line width=0.7pt](axis cs:0.236,1)circle(2.5pt);
\node[anchor=east,xshift=-3pt,font=\scriptsize,color=asrGray]at(axis cs:0.554,1){0.55};
\node[anchor=west,xshift=3pt,font=\scriptsize]at(axis cs:0.236,1){\textbf{0.24} ($-57.4\%$)};
\draw[black,line width=0.9pt,line cap=round,->,shorten >=3.5pt,shorten <=3.5pt]
  (axis cs:0.427,0)--(axis cs:0.178,0);
\filldraw[fill=white,draw=black,line width=0.7pt](axis cs:0.427,0)circle(2.5pt);
\filldraw[fill=black,draw=white,line width=0.7pt](axis cs:0.178,0)circle(2.5pt);
\node[anchor=east,xshift=-3pt,font=\scriptsize,color=asrGray]at(axis cs:0.427,0){0.43};
\node[anchor=west,xshift=3pt,font=\scriptsize]at(axis cs:0.178,0){\textbf{0.18} ($-58.3\%$)};
\draw[black,line width=0.9pt,line cap=round,->,shorten >=3.5pt,shorten <=3.5pt]
  (axis cs:0.328,-1)--(axis cs:0.196,-1);
\filldraw[fill=white,draw=black,line width=0.7pt](axis cs:0.328,-1)circle(2.5pt);
\filldraw[fill=black,draw=white,line width=0.7pt](axis cs:0.196,-1)circle(2.5pt);
\node[anchor=east,xshift=-3pt,font=\scriptsize,color=asrGray]at(axis cs:0.328,-1){0.33};
\node[anchor=west,xshift=3pt,font=\scriptsize]at(axis cs:0.196,-1){\textbf{0.20} ($-40.2\%$)};
\end{axis}
\end{tikzpicture}
\caption{ASR fine-tuning results on \textsc{WorldSpeech} with \texttt{whisper-large-v3-turbo}. For each target language, the open circle is the zero-shot baseline WER and the filled circle is the WER after fine-tuning on the \textsc{WorldSpeech} aligned-data split. WER can exceed 1.0 when the model produces more erroneous words than the reference contains, which occurs for zero-shot models on unseen languages. Evaluation is on the FLEURS test split where available, and on the \textsc{WorldSpeech} held-out test split for languages with no public benchmark (rows marked $^{\ast}$). Per-language WER and CER values are tabulated in Appendix~\ref{app:asr_results} (Table~\ref{tab:asr_results_v3}).}
\label{fig:asr_results}
\end{figure}

\paragraph{Setup.}
We fine-tune \texttt{whisper-large-v3-turbo}~\cite{radford2023whisper} on the \textsc{WorldSpeech} aligned-data split for each target language. The recipe is shared across all runs. We use AdamW~\cite{loshchilov2019adamw} with learning rate $10^{-5}$, effective batch size 32, bf16 mixed precision, linear warmup of 10\% of total steps capped at 500, and one pass over the training set. The Whisper \texttt{forced\_decoder\_ids} and \texttt{suppress\_tokens} masks are cleared during training and restored at evaluation with the target-language and \texttt{transcribe} task tokens. We evaluate with greedy decoding and a generation length of 225 tokens on the corresponding FLEURS~\cite{conneau2022fleurs} test split. For Albanian, Bahraini Arabic, Kreol Seselwa, Romansh, and Samoan no public benchmark with full coverage is available, so we evaluate on the \textsc{WorldSpeech} held-out test split (rows marked $^{\ast}$ in Figure~\ref{fig:asr_results} and Table~\ref{tab:asr_results_v3}). Training scripts and total compute expenditure can be found in Appendices~\ref{app:codebase} and~\ref{app:compute}, respectively.

\paragraph{Results.}
Fine-tuning on \textsc{WorldSpeech} improves the baseline across every language evaluated (Figure~\ref{fig:asr_results}). The largest gains concentrate on languages for which the baseline WER exceeds 1.0, where the fine-tuning contributes the bulk of the target-language signal: Samoan falls from 4.72 to 0.39 WER, Lao from 2.47 to 0.76, Romansh from 1.31 to 0.17, Georgian from 1.07 to 0.48, Burmese from 1.01 to 0.39, and Luxembourgish from 0.95 to 0.28. Even on languages where the baseline already produces a partially-correct transcript, fine-tuning roughly halves the error rate, as for Bahraini Arabic (0.62 to 0.30), Albanian (0.55 to 0.24), and Armenian (0.43 to 0.18). The full per-language WER and CER values are reported in Appendix~\ref{app:asr_results} (Table~\ref{tab:asr_results_v3}).

\paragraph{Tier thresholds.}

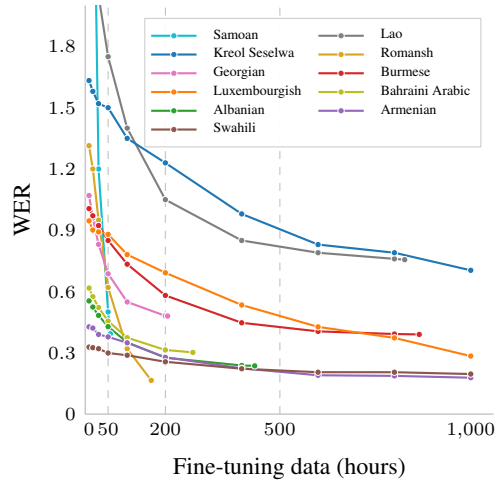
\begin{wrapfigure}{r}{0.5\textwidth}
\centering
\begin{tikzpicture}
\begin{axis}[
  width=0.5\textwidth,
  height=0.5\textwidth,
  xlabel={\small Fine-tuning data (hours)},
  ylabel={\small WER},
  xmin=-10, xmax=1060,
  ymin=0.0, ymax=2.0,
  xtick={0,50,200,500,1000},
  enlarge x limits=false,
  ytick={0,0.3,0.6,0.9,1.2,1.5,1.8},
  legend pos=north east,
  legend columns=2,
  legend style={font=\fontsize{5.25pt}{6pt}\selectfont, draw=gray!40, fill=white,
                fill opacity=0.94, text opacity=1,
                inner sep=2pt, row sep=-2pt, column sep=4pt},
  legend cell align=left,
  axis x line*=bottom,
  axis y line*=left,
  tick style={draw=none},
  axis line style={gray!60},
  label style={font=\small},
  tick label style={font=\scriptsize},
  grid=none,
]
\addplot[gray!50, dashed, thin, forget plot] coordinates {(50,0)(50,2.0)};
\addplot[gray!50, dashed, thin, forget plot] coordinates {(200,0)(200,2.0)};
\addplot[gray!50, dashed, thin, forget plot] coordinates {(500,0)(500,2.0)};
\addplot[colSamoan, thick, mark=*, mark size=1.2pt,
  mark options={fill=colSamoan, draw=white, line width=0.3pt}]
  coordinates {
  (0,4.719)(5,4.20)(10,3.10)(25,1.20)(50,0.50)(56,0.393)};
\addlegendentry{Samoan}
\addplot[colLao, thick, mark=*, mark size=1.2pt,
  mark options={fill=colLao, draw=white, line width=0.3pt}]
  coordinates {
  (0,2.469)(10,2.30)(25,2.05)(50,1.75)(100,1.40)(200,1.05)
  (400,0.85)(600,0.79)(800,0.76)(827,0.756)};
\addlegendentry{Lao}
\addplot[colSeychelles, thick, mark=*, mark size=1.2pt,
  mark options={fill=colSeychelles, draw=white, line width=0.3pt}]
  coordinates {
  (0,1.633)(10,1.580)(25,1.520)(50,1.500)(100,1.350)(200,1.230)
  (400,0.980)(600,0.830)(800,0.790)(1000,0.704)};
\addlegendentry{Kreol Seselwa}
\addplot[colRomansh, thick, mark=*, mark size=1.2pt,
  mark options={fill=colRomansh, draw=white, line width=0.3pt}]
  coordinates {
  (0,1.314)(10,1.20)(25,0.95)(50,0.62)(100,0.32)(163,0.165)};
\addlegendentry{Romansh}
\addplot[colGeorgian, thick, mark=*, mark size=1.2pt,
  mark options={fill=colGeorgian, draw=white, line width=0.3pt}]
  coordinates {
  (0,1.070)(10,0.960)(25,0.831)(50,0.687)(100,0.549)(200,0.481)(206,0.480)};
\addlegendentry{Georgian}
\addplot[colBurmese, thick, mark=*, mark size=1.2pt,
  mark options={fill=colBurmese, draw=white, line width=0.3pt}]
  coordinates {
  (0,1.006)(10,0.971)(25,0.923)(50,0.850)(100,0.734)(200,0.581)
  (400,0.447)(600,0.405)(800,0.392)(865,0.390)};
\addlegendentry{Burmese}
\addplot[colLuxembourgish, thick, mark=*, mark size=1.2pt,
  mark options={fill=colLuxembourgish, draw=white, line width=0.3pt}]
  coordinates {
  (0,0.946)(10,0.901)(25,0.892)(50,0.880)(100,0.781)(200,0.692)
  (400,0.534)(600,0.427)(800,0.373)(1000,0.284)};
\addlegendentry{Luxembourgish}
\addplot[colBahraini, thick, mark=*, mark size=1.2pt,
  mark options={fill=colBahraini, draw=white, line width=0.3pt}]
  coordinates {
  (0,0.617)(10,0.575)(25,0.522)(50,0.455)(100,0.374)(200,0.314)(272,0.302)};
\addlegendentry{Bahraini Arabic}
\addplot[colAlbanian, thick, mark=*, mark size=1.2pt,
  mark options={fill=colAlbanian, draw=white, line width=0.3pt}]
  coordinates {
  (0,0.554)(10,0.524)(25,0.483)(50,0.428)(100,0.351)(200,0.276)(400,0.238)(434,0.236)};
\addlegendentry{Albanian}
\addplot[colArmenian, thick, mark=*, mark size=1.2pt,
  mark options={fill=colArmenian, draw=white, line width=0.3pt}]
  coordinates {
  (0,0.427)(10,0.421)(25,0.390)(50,0.378)(100,0.349)(200,0.278)
  (400,0.224)(600,0.190)(800,0.187)(1000,0.178)};
\addlegendentry{Armenian}
\addplot[colSwahili, thick, mark=*, mark size=1.2pt,
  mark options={fill=colSwahili, draw=white, line width=0.3pt}]
  coordinates {
  (0,0.328)(10,0.325)(25,0.320)(50,0.299)(100,0.288)(200,0.256)
  (400,0.222)(600,0.205)(800,0.205)(1000,0.196)};
\addlegendentry{Swahili}
\end{axis}
\end{tikzpicture}
\caption{Hours-vs-WER ablation. Progressive fine-tuning of \texttt{whisper-large-v3-turbo} on hours-bounded subsamples of \textsc{WorldSpeech}, evaluated on FLEURS test (or the \textsc{WorldSpeech} held-out test for languages without FLEURS coverage). Each language begins from the baseline ASR ($x{=}0$) and its model is progressively trained on more hours. Sharp drop in WER occurs in the first 200h, with diminishing returns after 500h.}
\label{fig:hours_ablation}
\end{wrapfigure}

We define per-language data scale using four threshold tiers: 50, 200, 500, and 1{,}000 hours, matching the breakdown reported in the dataset comparison (Table~\ref{tab:dataset_comparison}). Figure~\ref{fig:hours_ablation} shows WER under progressive fine-tuning of \texttt{whisper-large-v3-turbo} on hours-bounded subsamples of \textsc{WorldSpeech} for eleven typologically distinct languages. WER decreases monotonically with available hours, with the steepest gains in the first 200 hours and diminishing returns past 500. The same shape holds for languages with exceedingly poor baselines (Samoan, Lao, Kreol Seselwa, Romansh, Georgian, Burmese, all with WER above 1), for languages with weak baselines (Luxembourgish, Bahraini Arabic, Albanian), and for languages where the pretrained model already produces partially-correct transcripts (Armenian, Swahili).

\section{Iterative Alignment Refinement}
\label{sec:iterative}

The number of hours extracted for a language depends on two factors: the quality of the human transcript paired with each recording, and the performance of the initial ASR on the target language. The first, collecting data, processing diverse formats and matching verbatim transcripts with corresponding audio is handled by our collection pipeline and matching strategy of Section~\ref{sec:pipeline}. Here we focus on improving the initial ASR. For low-resource languages, an initial model that has not seen (or very little) of the target language generates transcripts that diverge substantially from the human ground-truth transcript, and few segments pass the CER$<$0.3 filter. The languages that most need aligned data are therefore often those for which the pipeline yields the least data.

We address this by iterating the alignment process against the same set of human transcripts. The transcript paired with each recording stays fixed, and only the initial ASR changes between iterations. After a single pass over the full audio pool with the initial model, we fine-tune the initial ASR on the retained segments and repeat the alignment with this fine-tuned model. This improves the alignment yield of the second pass in two ways. First, as shown in Figure~\ref{fig:asr_results}, the fine-tuned models on \textsc{WorldSpeech} data significantly improve ASR performance on the target language. Second, the improved ASR has seen the target language in the specific speech setting of the source material (e.g. a trial testimony, an interview or a parliamentary speech). Therefore, its generated transcripts are more in-domain, recovering segments that the initial ASR could not match. This approach allows us to increase the number of aligned hours without additional data collection. A single iteration yields substantial gains across all languages tested, with relative improvements in retained hours ranging from $+19.5\%$ (Flemish) to $+201.1\%$ (Burmese), depending on how far the initial model's error rate is from the CER$<$0.3 alignment threshold.

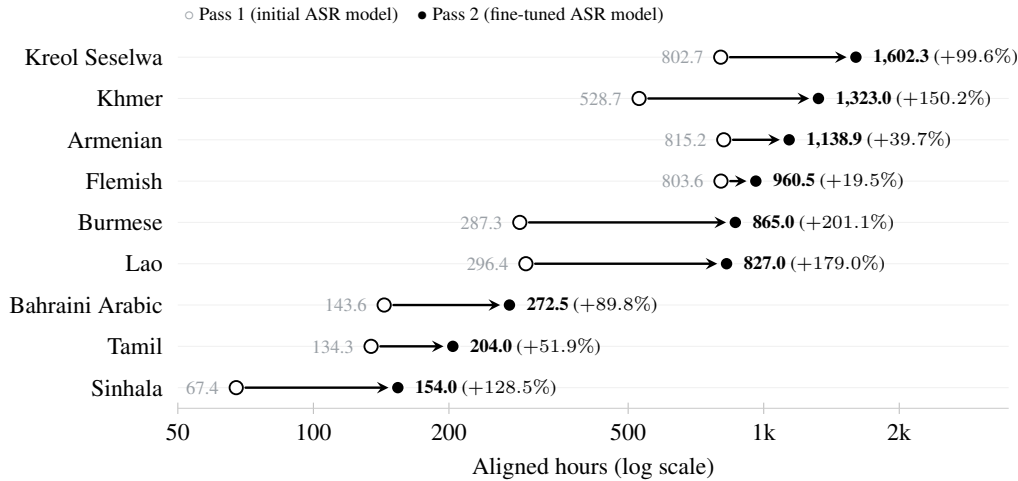
\begin{figure}[t!]
\centering
\definecolor{iarGray}{RGB}{154,160,166}
\begin{tikzpicture}[>=stealth]
\begin{axis}[
  xmode=log,
  xmin=50, xmax=3500,
  ymin=-0.5, ymax=8.5,
  ytick={0,1,2,3,4,5,6,7,8},
  yticklabels={
    {Sinhala},
    {Tamil},
    {Bahraini Arabic},
    {Lao},
    {Burmese},
    {Flemish},
    {Armenian},
    {Khmer},
    {Kreol Seselwa}
  },
  tick label style={font=\small},
  ytick style={draw=none},
  xtick={50,100,200,500,1000,2000},
  xticklabels={50,100,200,500,1k,2k},
  xtick style={color=gray!50,thin},
  xlabel={Aligned hours (log scale)},
  xlabel style={font=\small},
  axis x line=bottom,
  axis y line=left,
  x axis line style={color=gray!40, -},
  y axis line style={draw=none},
  ymajorgrids=true,
  grid style={gray!12,very thin},
  width=0.9\textwidth,
  height=6.5cm,
  clip=false,
]
\draw[black,line width=0.9pt,line cap=round,->,shorten >=3.5pt,shorten <=3.5pt]
  (axis cs:802.7,8)--(axis cs:1602.3,8);
\filldraw[fill=white,draw=black,line width=0.7pt](axis cs:802.7,8)circle(2.5pt);
\filldraw[fill=black,draw=white,line width=0.7pt](axis cs:1602.3,8)circle(2.5pt);
\node[anchor=east,xshift=-3pt,font=\scriptsize,color=iarGray]at(axis cs:802.7,8){802.7};
\node[anchor=west,xshift=3pt,font=\scriptsize]at(axis cs:1602.3,8){\textbf{1{,}602.3} ($+99.6\%$)};
\draw[black,line width=0.9pt,line cap=round,->,shorten >=3.5pt,shorten <=3.5pt]
  (axis cs:528.7,7)--(axis cs:1323.0,7);
\filldraw[fill=white,draw=black,line width=0.7pt](axis cs:528.7,7)circle(2.5pt);
\filldraw[fill=black,draw=white,line width=0.7pt](axis cs:1323.0,7)circle(2.5pt);
\node[anchor=east,xshift=-3pt,font=\scriptsize,color=iarGray]at(axis cs:528.7,7){528.7};
\node[anchor=west,xshift=3pt,font=\scriptsize]at(axis cs:1323.0,7){\textbf{1{,}323.0} ($+150.2\%$)};
\draw[black,line width=0.9pt,line cap=round,->,shorten >=3.5pt,shorten <=3.5pt]
  (axis cs:815.2,6)--(axis cs:1138.9,6);
\filldraw[fill=white,draw=black,line width=0.7pt](axis cs:815.2,6)circle(2.5pt);
\filldraw[fill=black,draw=white,line width=0.7pt](axis cs:1138.9,6)circle(2.5pt);
\node[anchor=east,xshift=-3pt,font=\scriptsize,color=iarGray]at(axis cs:815.2,6){815.2};
\node[anchor=west,xshift=3pt,font=\scriptsize]at(axis cs:1138.9,6){\textbf{1{,}138.9} ($+39.7\%$)};
\draw[black,line width=0.9pt,line cap=round,->,shorten >=3.5pt,shorten <=3.5pt]
  (axis cs:803.6,5)--(axis cs:960.5,5);
\filldraw[fill=white,draw=black,line width=0.7pt](axis cs:803.6,5)circle(2.5pt);
\filldraw[fill=black,draw=white,line width=0.7pt](axis cs:960.5,5)circle(2.5pt);
\node[anchor=east,xshift=-3pt,font=\scriptsize,color=iarGray]at(axis cs:803.6,5){803.6};
\node[anchor=west,xshift=3pt,font=\scriptsize]at(axis cs:960.5,5){\textbf{960.5} ($+19.5\%$)};
\draw[black,line width=0.9pt,line cap=round,->,shorten >=3.5pt,shorten <=3.5pt]
  (axis cs:287.3,4)--(axis cs:865.0,4);
\filldraw[fill=white,draw=black,line width=0.7pt](axis cs:287.3,4)circle(2.5pt);
\filldraw[fill=black,draw=white,line width=0.7pt](axis cs:865.0,4)circle(2.5pt);
\node[anchor=east,xshift=-3pt,font=\scriptsize,color=iarGray]at(axis cs:287.3,4){287.3};
\node[anchor=west,xshift=3pt,font=\scriptsize]at(axis cs:865.0,4){\textbf{865.0} ($+201.1\%$)};
\draw[black,line width=0.9pt,line cap=round,->,shorten >=3.5pt,shorten <=3.5pt]
  (axis cs:296.4,3)--(axis cs:827.0,3);
\filldraw[fill=white,draw=black,line width=0.7pt](axis cs:296.4,3)circle(2.5pt);
\filldraw[fill=black,draw=white,line width=0.7pt](axis cs:827.0,3)circle(2.5pt);
\node[anchor=east,xshift=-3pt,font=\scriptsize,color=iarGray]at(axis cs:296.4,3){296.4};
\node[anchor=west,xshift=3pt,font=\scriptsize]at(axis cs:827.0,3){\textbf{827.0} ($+179.0\%$)};
\draw[black,line width=0.9pt,line cap=round,->,shorten >=3.5pt,shorten <=3.5pt]
  (axis cs:143.6,2)--(axis cs:272.5,2);
\filldraw[fill=white,draw=black,line width=0.7pt](axis cs:143.6,2)circle(2.5pt);
\filldraw[fill=black,draw=white,line width=0.7pt](axis cs:272.5,2)circle(2.5pt);
\node[anchor=east,xshift=-3pt,font=\scriptsize,color=iarGray]at(axis cs:143.6,2){143.6};
\node[anchor=west,xshift=3pt,font=\scriptsize]at(axis cs:272.5,2){\textbf{272.5} ($+89.8\%$)};
\draw[black,line width=0.9pt,line cap=round,->,shorten >=3.5pt,shorten <=3.5pt]
  (axis cs:134.3,1)--(axis cs:204.0,1);
\filldraw[fill=white,draw=black,line width=0.7pt](axis cs:134.3,1)circle(2.5pt);
\filldraw[fill=black,draw=white,line width=0.7pt](axis cs:204.0,1)circle(2.5pt);
\node[anchor=east,xshift=-3pt,font=\scriptsize,color=iarGray]at(axis cs:134.3,1){134.3};
\node[anchor=west,xshift=3pt,font=\scriptsize]at(axis cs:204.0,1){\textbf{204.0} ($+51.9\%$)};
\draw[black,line width=0.9pt,line cap=round,->,shorten >=3.5pt,shorten <=3.5pt]
  (axis cs:67.4,0)--(axis cs:154.0,0);
\filldraw[fill=white,draw=black,line width=0.7pt](axis cs:67.4,0)circle(2.5pt);
\filldraw[fill=black,draw=white,line width=0.7pt](axis cs:154.0,0)circle(2.5pt);
\node[anchor=east,xshift=-3pt,font=\scriptsize,color=iarGray]at(axis cs:67.4,0){67.4};
\node[anchor=west,xshift=3pt,font=\scriptsize]at(axis cs:154.0,0){\textbf{154.0} ($+128.5\%$)};
\node[anchor=south west,font=\scriptsize,inner sep=2pt]
  at (rel axis cs:0,1.02) {%
    \textcolor{iarGray}{$\circ$}~Pass 1 (initial ASR model)\quad
    $\bullet$~Pass 2 (fine-tuned ASR model)};
\end{axis}
\end{tikzpicture}
\caption{Aligned hours after one iteration of iterative alignment refinement. Each bar is the pass-2 (fine-tuned) total: the gray segment is the number of hours of aligned data that was retained in pass 1 with the initial ASR, and the blue segment is the additional hours recovered by the language-adapted model in pass 2. The percentage at the right of each bar is the relative gain over pass 1. We find that gains scale inversely with the initial model's quality on each target language.}
\label{fig:iar_dumbbell}
\end{figure}

Figure~\ref{fig:iar_dumbbell} shows pass-1 and pass-2 aligned hours for nine languages spanning diverse scripts and geographic regions. Gains are smallest ($+19.5\%$ for Flemish) for languages where the initial model already produces reasonable transcripts, and largest ($+150.2\%$ for Khmer, $+179.0\%$ for Lao, $+201.1\%$ for Burmese) where the initial ASR model performs poorly on non-Latin scripts. We experimented with additional passes, but found that a third pass adds only $+0.2\%$ to $+8.8\%$ further hours over Pass 2 (average $+4.3\%$ versus $+95.4\%$ for Pass 2 over Pass 1). The per-language Pass 3 yield is reported in Appendix~\ref{app:pass3}.

\section{Limitations}
\label{sec:limitations}

\textsc{WorldSpeech} inherits the bias of its sources. Most aligned hours come from parliamentary debates and broadcast news, with smaller contributions from public-domain audiobooks. Speakers are skewed towards adult, formally educated and from public-facing professions, and they are not demographically representative of the languages they speak. The speaking style is closer to formal prepared speech than to spontaneous conversation. We diversify this in part by including non-parliamentary sources spanning broadcast news (RFA, VOA, RFE/RL) and read literary speech (LibriVox~\cite{librivox}, Aozora~\cite{aozora}, Ben-Yehuda~\cite{benyehuda}), but the overall distribution remains biased toward formal speech and downstream models trained on \textsc{WorldSpeech} alone may underperform on conversational input. Additionally, alignment quality is bounded by the off-the-shelf ASR used during alignment, which varies substantially by language. For languages where Whisper~\cite{radford2023whisper} or MMS~\cite{pratap2024mms} transcribe poorly, fewer segments pass the CER$<$0.3 filter and the per-language yield is correspondingly lower. We counteract this issue to some extent using the iterative alignment refinement to obtain more segments.

\section{Conclusion}
In this work, we introduced \textsc{WorldSpeech}, a multilingual speech corpus of 76 languages totaling 65k ground-truth aligned hours, sourced from public archives. We assembled the corpus by extending existing long-audio alignment pipelines with source-specific collection and preprocessing approaches, and leveraged an iterative alignment refinement procedure that fine-tuned an initial ASR on the first-pass yield and re-aligned parts of the remaining unaligned utterances, recovering between $+19.5\%$ and $+201.1\%$ additional aligned hours per language. Among publicly available ground-truth aligned multilingual corpora, \textsc{WorldSpeech} covers more languages than prior publicly available work, with 53 languages above 50 hours, 37 above 200 hours, 28 above 500 hours, and 24 above 1k hours of human-transcribed audio. Fine-tuning \texttt{whisper-large-v3-turbo} on 11 typologically diverse languages reduces WER on average by $63.5\%$ relative to the zero-shot baseline, including languages where the zero-shot model produces more erroneous than reference words. \textsc{WorldSpeech} provides a public speech corpus training resource at the scale required for substantially improved ASR on multiple previously underserved languages.


\bibliographystyle{plainnat}
\bibliography{references}

\clearpage
\appendix
\section*{Appendix}

\section{Per-language alignment ASR}
\label{app:asr_alignment_models}
For each configuration the alignment ASR model is selected by ablating before committing to the full run. Two to three candidate models are evaluated on approximately 10 hours of the available audio and the model that maximises the fraction of segments passing the CER$<$0.3 threshold is chosen. In practice this means whisper-large-v3-turbo suffices for most European and widely-resourced languages, MMS-1B with its per-language adapter is better suited for languages where Whisper produces script errors or high hallucination rates, and a community fine-tune is used when one exists that was trained specifically on the target dialect. Table~\ref{tab:asr_alignment_models} reports the chosen model per language, sorted alphabetically.

\begin{small}
\setlength{\tabcolsep}{4pt}
\renewcommand{\arraystretch}{0.95}
\begin{longtable}{@{}p{3.8cm} p{2.2cm} p{6.8cm}@{}}
\caption{Alignment ASR model chosen per language via the 10-hour ablation, sorted alphabetically by language.}
\label{tab:asr_alignment_models} \\
\toprule
Language & Config & Alignment ASR \\
\midrule
\endfirsthead
\multicolumn{3}{l}{\emph{(Table~\ref{tab:asr_alignment_models} continued)}} \\
\toprule
Language & Config & Alignment ASR \\
\midrule
\endhead
\midrule
\multicolumn{3}{r}{\emph{(continued on next page)}} \\
\endfoot
\bottomrule
\endlastfoot
Albanian                        & al\_sq            & openai/whisper-large-v3-turbo \\
Albanian (Kosovo)               & xk\_sq            & openai/whisper-large-v3-turbo \\
Amharic (LibriVox)             & am                & faster-whisper large-v3 \\
Amharic (VOA)                  & et\_am\_voa       & facebook/mms-1b-all (\texttt{amh}) \\
Ancient Greek                   & gr\_grc           & facebook/mms-1b-all (\texttt{grc}) \\
Arabic (Algeria)                & dz\_ar            & faster-whisper large-v3 \\
Arabic (Bahrain)                & bh\_ar\_nuwab     & faster-whisper large-v3 \\
Arabic (Egypt)                  & eg\_ar            & faster-whisper large-v3 \\
Arabic (Iraq)                   & iq\_ar            & faster-whisper large-v3 \\
Arabic (Kuwait)                 & kw\_ar            & faster-whisper large-v3 \\
Arabic (Morocco)                & ma\_ar            & faster-whisper large-v3 \\
Arabic (Oman)                   & om\_ar            & faster-whisper large-v3 \\
Arabic (Saudi Arabia)           & sa\_ar            & faster-whisper large-v3 \\
Arabic (Tunisia)                & tn\_ar            & faster-whisper large-v3 \\
Arabic (UN)                     & un\_ar            & faster-whisper large-v3 \\
Armenian                        & am\_hy            & facebook/mms-1b-all (\texttt{hye}) \\
Azerbaijani                     & az\_az            & faster-whisper large-v3 \\
Bambara                         & ml\_bm            & facebook/mms-1b-all (\texttt{bam}) \\
Belarusian                      & by\_be            & openai/whisper-large-v3 \\
Bengali                         & bd\_bn            & facebook/mms-1b-all (\texttt{ben}) \\
Burmese                         & mm\_my            & facebook/mms-1b-all (\texttt{mya}) \\
Burmese (RFA)                   & mm\_my\_rfa       & facebook/mms-1b-all (\texttt{mya}) \\
Cantonese (HK committees)       & hk\_yue (comm.)  & openai/whisper-large-v3-turbo \\
Cantonese (HK plenary)          & hk\_yue           & openai/whisper-large-v3 \\
Cantonese (HK, RFA)             & hk\_yue\_rfa      & faster-whisper large-v3 \\
Cantonese (Macau)               & mo\_yue           & openai/whisper-large-v3-turbo \\
Central Kurdish (Sorani)        & krd\_ckb          & kurdai-academy/mms-asr-1b-ckb-v4 \\
Czech                           & cz\_cs            & openai/whisper-large-v3-turbo \\
Dhivehi                         & mv\_dv            & facebook/mms-1b-all (\texttt{div}) \\
Dutch (Belgium)                 & be\_nl            & openai/whisper-large-v3-turbo \\
Dutch (Netherlands)             & nl                & openai/whisper-large-v3-turbo \\
English (Australia)             & au\_en            & faster-whisper large-v3 \\
English (ICC)                   & icc               & openai/whisper-large-v3-turbo \\
English (Jamaica)               & jm\_en            & openai/whisper-large-v3-turbo \\
English (Kenya)                 & ke\_en            & openai/whisper-large-v3-turbo \\
English (New Zealand)           & nz\_en            & faster-whisper large-v3 \\
English (Scotland)              & gb\_sco           & openai/whisper-large-v3-turbo \\
English (Sierra Leone)          & sl\_en            & openai/whisper-large-v3-turbo \\
English (Zambia)                & zm\_en            & openai/whisper-large-v3-turbo \\
Esperanto                       & xx\_eo            & openai/whisper-large-v3-turbo \\
Faroese                         & fo\_fo            & facebook/mms-1b-all (\texttt{fao}) \\
Fijian                          & fj\_fj            & facebook/mms-1b-all (\texttt{fij}) \\
French (C\^ote d'Ivoire, ICC)   & ci\_fr            & openai/whisper-large-v3-turbo \\
French (DRC, ICC)               & cd                & openai/whisper-large-v3-turbo \\
French (Qu\'ebec)               & ca\_fr            & openai/whisper-large-v3-turbo \\
Fula                            & sn\_ff            & facebook/mms-1b-all (\texttt{ful}) \\
Galician                        & gl                & openai/whisper-large-v3-turbo \\
Georgian                        & ge\_ka            & openai/whisper-large-v3 \\
German (Austria)                & at\_de            & openai/whisper-large-v3-turbo \\
Gilbertese                      & ki\_gil           & facebook/mms-1b-all (\texttt{gil}) \\
Greek                           & gr\_el            & openai/whisper-large-v3-turbo \\
Greek (Cyprus)                  & cy\_el            & openai/whisper-large-v3-turbo \\
Hausa (Chad)                    & td\_ha            & facebook/mms-1b-all (\texttt{hau}) \\
Hausa (Nigeria)                 & ng\_ha            & facebook/mms-1b-all (\texttt{hau}) \\
Hebrew                          & il\_he            & ivrit-ai/whisper-large-v3-turbo-ct2 \\
Hindi (Bihar)                   & in\_hi\_bh        & openai/whisper-large-v3 \\
Hindi (Chhattisgarh)            & in\_hi\_cg        & facebook/mms-1b-all (\texttt{hin}) \\
Hindi (Mann Ki Baat)            & in\_hi\_mkb       & facebook/mms-1b-all (\texttt{hin}) \\
Hindi (national)                & in\_hi            & faster-whisper large-v3 \\
Hindi (Rajasthan)               & in\_hi\_rj        & openai/whisper-large-v3 \\
Hungarian                       & hu                & openai/whisper-large-v3-turbo \\
Igbo                            & ng\_ig            & facebook/mms-1b-all (\texttt{ibo}) \\
Indonesian                      & id\_id            & facebook/mms-1b-all (\texttt{ind}) \\
Inuktitut                       & ca\_iu            & openai/whisper-large-v3-turbo \\
Irish                           & ga (MMS)          & facebook/mms-1b-all (\texttt{gle}) \\
Irish                           & ga (Whisper)      & faster-whisper large-v3 \\
Japanese                        & jp\_ja            & kotoba-tech/kotoba-whisper-v1.0-faster \\
Kazakh                          & kz\_kk            & openai/whisper-large-v3 \\
Khmer                           & kh\_km            & facebook/mms-1b-all (\texttt{khm}) \\
Kinyarwanda                     & rw\_rw            & facebook/mms-1b-all (\texttt{kin}) \\
Korean                          & kr\_ko            & faster-whisper large-v3 \\
Kreol Seselwa                   & sc\_crs           & facebook/mms-1b-all (\texttt{crs}) \\
Kurdish (Kurmanji)              & voa\_kmr          & facebook/mms-1b-all (\texttt{kmr}) \\
Kyrgyz                          & kg\_ky            & facebook/mms-1b-all (\texttt{kir}) \\
Lao                             & la\_lo            & facebook/mms-1b-all (\texttt{lao}) \\
Luxembourgish                   & lu\_lb            & in-house whisper-lu\_lb-v2 \\
Malay                           & my\_ms            & faster-whisper large-v3 \\
Malayalam                       & in\_ml            & facebook/mms-1b-all (\texttt{mal}) \\
M\=aori                         & nz\_mi            & faster-whisper large-v3 \\
Marathi                         & in\_mr            & facebook/mms-1b-all (\texttt{mar}) \\
Mandarin (Taiwan)               & tw\_zh            & openai/whisper-large-v3-turbo \\
Mongolian                       & mn\_mn            & bayartsogt/whisper-large-v2-mn-13 \\
Montenegrin                     & me\_cnr           & openai/whisper-large-v3-turbo \\
Morisyen                        & mu\_mfe           & facebook/mms-1b-all (\texttt{mfe}) \\
Ndebele                         & zw\_nd            & openai/whisper-large-v3-turbo \\
Nepali                          & np\_ne            & Dragneel/whisper-medium-nepali-ct2 \\
Oromo                           & et\_om            & facebook/mms-1b-all (\texttt{orm}) \\
Papiamentu                      & cw\_pap           & facebook/mms-1b-all (\texttt{pap}) \\
Pashto (Afghanistan)            & af\_ps            & facebook/mms-1b-all (\texttt{pus}) \\
Pashto (Pakistan)               & pk\_ps            & facebook/mms-1b-all (\texttt{pus}) \\
Persian (Afghanistan)           & af\_fa            & facebook/mms-1b-all (\texttt{fas}) \\
Persian (Iran)                  & ir\_fa            & facebook/mms-1b-all (\texttt{fas}) \\
Polish                          & pl\_pl            & openai/whisper-large-v3 \\
Portuguese (Brazil, C\^amara)   & br\_pt\_camara    & distil-whisper/distil-large-v3 \\
Portuguese (Brazil, Senado)     & br\_pt            & openai/whisper-large-v3-turbo \\
Punjabi                         & in\_pa            & facebook/mms-1b-all (\texttt{pan}) \\
Romanian                        & ro\_ro            & openai/whisper-large-v3-turbo \\
Romanian (Moldova)              & md\_ro            & openai/whisper-large-v3-turbo \\
Romansh                         & ch\_rm            & infinitejoy/wav2vec2-large-xls-r-300m-rm \\
Russian                         & ru\_ru            & openai/whisper-large-v3-turbo \\
Russian (Belarus)               & by\_ru            & openai/whisper-large-v3 \\
Samoan                          & ws\_sm            & facebook/mms-1b-all (\texttt{smo}) \\
Shona                           & zw\_sn            & openai/whisper-large-v3-turbo \\
Somali                          & sl\_so            & facebook/mms-1b-all (\texttt{som}) \\
Somali                          & so\_so            & facebook/mms-1b-all (\texttt{som}) \\
Sotho (Lesotho)                 & ls\_st            & faster-whisper large-v3 \\
South African languages         & za\_parliament    & guymandude/MMS-ASR-South-African-11 \\
Spanish (Argentina)             & ar\_es            & openai/whisper-large-v3-turbo \\
Spanish (Chile)                 & cl\_es            & openai/whisper-large-v3-turbo \\
Spanish (Colombia)              & co\_es            & openai/whisper-large-v3-turbo \\
Spanish (Mexico)                & mx\_es            & openai/whisper-large-v3-turbo \\
Spanish (Paraguay)              & py\_es            & openai/whisper-large-v3-turbo \\
Spanish (Peru)                  & pe\_es            & openai/whisper-large-v3-turbo \\
Spanish (Puerto Rico)           & pr\_es            & openai/whisper-large-v3-turbo \\
Spanish (Uruguay)               & uy\_es            & openai/whisper-large-v3-turbo \\
Swahili (Tanzania)              & tz\_sw            & faster-whisper large-v3 \\
Swahili (Zanzibar)              & tz\_zw            & faster-whisper large-v3 \\
Swedish (\r{A}land)             & ax\_sv            & openai/whisper-large-v3-turbo \\
Tagalog                         & ph\_tl            & openai/whisper-large-v3 \\
Tajik                           & tj\_tg            & facebook/mms-1b-all (\texttt{tgk}) \\
Tamil (Sri Lanka)               & lk\_ta            & facebook/mms-1b-all (\texttt{tam}) \\
Thai                            & th\_th            & facebook/mms-1b-all (\texttt{tha}) \\
Tibetan                         & cn\_bo            & facebook/mms-1b-all (\texttt{bod}) \\
Tibetan (RFA)                   & cn\_bo\_rfa       & facebook/mms-1b-all (\texttt{bod}) \\
Tigrinya                        & et\_ti            & facebook/mms-1b-all (\texttt{tir}) \\
Tswana (Botswana)               & bw\_parliament    & facebook/mms-1b-all (\texttt{tsn}) \\
Turkish                         & tr                & openai/whisper-large-v3-turbo \\
Turkmen                         & tm\_tk            & facebook/mms-1b-all (\texttt{tuk}) \\
Urdu                            & pk\_ur            & facebook/mms-1b-all (\texttt{urd}) \\
Uyghur                          & cn\_ug            & facebook/mms-1b-all (\texttt{uig}) \\
Uzbek                           & uz\_uz            & facebook/mms-1b-all (\texttt{uzb}) \\
Vietnamese                      & vn\_vi            & facebook/mms-1b-all (\texttt{vie}) \\
\end{longtable}
\end{small}

\section{Sources per country-language configuration}
\label{app:sources}
Table~\ref{tab:sources} lists the source or sources we used for each row of the main per-language overview (Figure~\ref{fig:dataset_overview}). Several configurations merge multiple sources for the same country-language pair, in which case all of them are listed in a single row.

\begin{small}
\setlength{\tabcolsep}{4pt}
\renewcommand{\arraystretch}{0.95}
\begin{longtable}{@{}p{0.24\textwidth} p{0.18\textwidth} p{0.55\textwidth}@{}}
\caption{Sources for each country-language configuration in \textsc{WorldSpeech}, in the same hours-descending order as Figure~\ref{fig:dataset_overview}.}
\label{tab:sources} \\
\toprule
Country & Language & Source(s) \\
\midrule
\endfirsthead
\multicolumn{3}{l}{\emph{(Table~\ref{tab:sources} continued)}} \\
\toprule
Country & Language & Source(s) \\
\midrule
\endhead
\midrule
\multicolumn{3}{r}{\emph{(continued on next page)}} \\
\endfoot
\bottomrule
\endlastfoot
Hong Kong                        & Cantonese          & Legislative Council \\
Chile                            & Spanish            & Chamber of Deputies and Senate \\
Seychelles                       & Kreol Seselwa      & National Assembly \\
Russia                           & Russian            & State Duma \\
Japan                            & Japanese           & LibriVox audiobooks and Aozora Bunko readings \\
Cambodia                         & Khmer              & Radio Free Asia, Khmer Service \\
Canada (Quebec)                  & French             & Quebec National Assembly \\
Austria                          & German             & National Council and Federal Council \\
Moldova                          & Romanian           & Parliament of Moldova, privesc.eu HLS streams and stenogram PDFs \\
Belgium                          & Dutch              & Flemish Parliament \\
Brazil                           & Portuguese         & Federal Senate \\
Uruguay                          & Spanish            & Chamber of Representatives and Senate \\
Armenia                          & Armenian           & National Assembly \\
India (Rajasthan)                & Hindi              & Rajasthan Vidhan Sabha \\
Myanmar                          & Burmese            & Pyidaungsu Hluttaw \\
Laos                             & Lao                & Radio Free Asia, Lao Service \\
Mexico                           & Spanish            & Mexico City Congress and Supreme Court of Justice of the Nation \\
Vietnam                          & Vietnamese         & Radio Free Asia, Vietnamese Service \\
Tanzania                         & Swahili            & Bunge of the United Republic of Tanzania \\
Romania                          & Romanian           & Senate of Romania \\
Hungary                          & Hungarian          & National Assembly \\
Australia                        & English            & House of Representatives and Senate, Australian Parliament House \\
South Korea                      & Korean             & National Assembly plenary and committee sessions \\
Taiwan                           & Mandarin           & Legislative Yuan IVOD \\
Cyprus                           & Greek              & House of Representatives \\
Azerbaijan                       & Azerbaijani        & Voice of America, Azerbaijani Service \\
Malaysia                         & Malay              & Parliament of Malaysia \\
Luxembourg                       & Luxembourgish      & Chamber of Deputies \\
Zambia                           & English            & National Assembly of Zambia \\
Albania                          & Albanian           & Assembly of the Republic of Albania, with additional public-domain Albanian recordings \\
Argentina                        & Spanish            & Chamber of Deputies and Senate \\
New Zealand                      & English            & House of Representatives \\
India (Bihar)                    & Hindi              & Bihar Vidhan Sabha \\
Philippines                      & Tagalog            & House of Representatives and Senate \\
Bahrain                          & Arabic             & Council of Representatives \\
Georgia                          & Georgian           & Parliament of Georgia \\
China (Xinjiang)                 & Uyghur             & Radio Free Asia, Uyghur Service \\
Puerto Rico                      & Spanish            & House of Representatives and Senate \\
Mongolia                         & Mongolian          & State Great Khural and Latter-day Saints addresses \\
Kazakhstan                       & Kazakh             & Mazhilis \\
Kosovo                           & Albanian           & Assembly of the Republic of Kosovo \\
Switzerland                      & Romansh            & Radio Televisiun Svizra Rumantscha \\
Kenya                            & English            & Parliament of Kenya \\
Sri Lanka                        & Sinhala            & Parliament of Sri Lanka \\
Colombia                         & Spanish            & House of Representatives and Senate \\
Paraguay                         & Spanish            & Chamber of Deputies and Senate \\
Sierra Leone                     & English            & Parliament of Sierra Leone \\
Iraq                             & Arabic             & Council of Representatives \\
Indonesia                        & Indonesian         & Voice of America, Indonesian Service \\
Algeria                          & Arabic             & National People's Assembly and Journal Officiel des D\'ebats \\
Morocco                          & Arabic             & House of Representatives and House of Councillors \\
\r{A}land Islands                & Swedish            & Lagting \\
Ireland                          & Irish              & Houses of the Oireachtas \\
Nepal                            & Nepali             & Federal Parliament of Nepal \\
Samoa                            & Samoan             & Legislative Assembly of Samoa \\
Nigeria                          & Hausa              & Voice of America, Hausa Service \\
Botswana                         & Tswana             & National Assembly of Botswana \\
Montenegro                       & Montenegrin        & Parliament of Montenegro \\
Bangladesh                       & Bengali            & Jatiya Sangsad \\
India (national)                 & Hindi              & Mann Ki Baat national radio address \\
Israel                           & Hebrew             & LibriVox audiobooks and Ben-Yehuda Project \\
Ethiopia                         & Amharic            & Voice of America, Amharic Service \\
Mauritius                        & Morisyen           & National Assembly of Mauritius \\
Greece                           & Greek              & Hellenic Parliament \\
Iraqi Kurdistan                  & Central Kurdish    & Kurdistan Parliament \\
Uzbekistan                       & Uzbek              & Ozodlik, Radio Free Europe / Radio Liberty \\
Nigeria                          & Igbo               & Voice of America, Igbo Service \\
Canada (Nunavut)                 & Inuktitut          & Legislative Assembly of Nunavut \\
Iran                             & Persian            & Voice of America, Persian Service \\
Democratic Republic of the Congo & French             & International Criminal Court trials of Lubanga, Ntaganda, Bemba, Katanga and Chui \\
Belarus                          & Belarusian         & Knihi.com Belarusian audiobook archive \\
Egypt                            & Arabic             & House of Representatives and State Information Service \\
Maldives                         & Dhivehi            & People's Majlis \\
Zimbabwe                         & Shona              & Voice of America, Shona Service \\
Rwanda                           & Kinyarwanda        & Chamber of Deputies \\
International                    & Esperanto          & LibriVox audio-books \\
Eritrea                          & Tigrinya           & Voice of America, Tigrinya Service \\
Tanzania (Zanzibar)              & Swahili            & Zanzibar House of Representatives \\
C\^ote d'Ivoire                  & French             & International Criminal Court trial of Gbagbo and Bl\'e Goud\'e \\
United Nations                   & Arabic             & UN General Assembly and Security Council sessions \\
Ethiopia                         & Oromo              & Voice of America, Oromo Service \\
South Africa                     & Afrikaans          & Parliament of South Africa \\
South Africa                     & Zulu               & Parliament of South Africa \\
Jamaica                          & English            & Parliament of Jamaica \\
Saudi Arabia                     & Arabic             & Public-domain Arabic audio (Internet Archive) \\
South Africa                     & Xhosa              & Parliament of South Africa \\
India (Kerala)                   & Malayalam          & Kerala Legislative Assembly \\
India (Punjab)                   & Punjabi            & Punjab Vidhan Sabha \\
Belarus                          & Russian            & Lukashenko Poslanie presidential addresses \\
Greece (classical)               & Ancient Greek      & LibriVox readings \\
India (Maharashtra)              & Marathi            & Maharashtra Vidhan Sabha \\
New Zealand                      & M\=aori            & House of Representatives, te reo passages of bound bilingual sittings \\
South Africa                     & Northern Sotho     & Parliament of South Africa \\
South Africa                     & Tsonga             & Parliament of South Africa \\
Hong Kong                        & Cantonese          & Radio Free Asia, Cantonese Service \\
\end{longtable}
\end{small}

\section{Prior open-source aligned data per language}
\label{app:prior_census}

Table~\ref{tab:sota_prior} reports, for each \textsc{WorldSpeech} language.  Single largest publicly redistributable ground-truth aligned corpus that existed before \textsc{WorldSpeech}. Hours are taken directly from the cited primary source (paper, dataset card, or official release page). Auto-generated and pseudo-labelled corpora are excluded (GigaSpeech~2, YODAS, Emilia, ReazonSpeech, MSR-86k auto portions). Corpora requiring institutional access or restricted to a single country's residents are excluded (KsponSpeech, BEA database, CGN). Common Voice hours are validated hours from v25.0 (2026-03-09)~\cite{ardila2020common}. FLEURS hours are the per-language training split ($\sim$10\,h) from~\cite{conneau2022fleurs}. Languages in italics are those where a larger prior corpus already existed; all others represent cases where \textsc{WorldSpeech} is the largest or first public ground-truth resource.

\begin{small}
\setlength{\tabcolsep}{3pt}
\renewcommand{\arraystretch}{0.95}
\begin{longtable}{@{}
  p{0.20\linewidth}
  p{0.05\linewidth}
  p{0.43\linewidth}
  >{\raggedleft\arraybackslash}p{0.08\linewidth}
  >{\raggedleft\arraybackslash}p{0.10\linewidth}
  >{\raggedleft\arraybackslash}p{0.10\linewidth}
@{}}
\caption{Largest prior publicly redistributable ground-truth aligned corpus per language vs.\ \textsc{WorldSpeech}, sorted by \textsc{WorldSpeech} hours descending. \textbf{New} = no prior public corpus identified. Italic language names indicate cases where the prior corpus is larger than \textsc{WorldSpeech}. All hours verified from primary sources; see text for exclusion criteria.}
\label{tab:sota_prior} \\
\toprule
Language & ISO & Prior corpus & Prior\,h & WorldSpeech\,h & $\times$ \\
\midrule
\endfirsthead
\multicolumn{6}{l}{\emph{(Table~\ref{tab:sota_prior} continued)}} \\
\toprule
Language & ISO & Prior corpus & Prior\,h & WS\,h & $\times$ \\
\midrule
\endhead
\midrule
\multicolumn{6}{r}{\emph{(continued on next page)}} \\
\endfoot
\bottomrule
\endlastfoot
Dutch (Flemish)                     & nl-BE & {--}                                                    & {--}         & $\mathbf{961}$     & \textbf{new} \\
Burmese                             & my    & FLEURS~\cite{conneau2022fleurs}                         & $18$         & $\mathbf{865}$     & $\mathbf{48}{\times}$ \\
Lao                                 & lo    & FLEURS~\cite{conneau2022fleurs}                         & $10$         & $\mathbf{827}$     & $\mathbf{83}{\times}$ \\
Vietnamese                          & vi    & VIVOS                                                   & $15$         & $\mathbf{726}$     & $\mathbf{48}{\times}$ \\
Albanian                            & sq    & Common Voice 25~\cite{ardila2020common}                 & $9$          & $\mathbf{434}$     & $\mathbf{48}{\times}$ \\
Malay                               & ms    & FLEURS~\cite{conneau2022fleurs}                         & $10$         & $\mathbf{432}$     & $\mathbf{43}{\times}$ \\
\textit{Greek}                      & el    & EuroSpeech~\cite{pfisterer2025eurospeech}               & $2{,}395$    & $430$              & $0.18{\times}$ \\
Indonesian                          & id    & Common Voice 25~\cite{ardila2020common}                 & $34$         & $\mathbf{340}$     & $\mathbf{10}{\times}$ \\
Azerbaijani                         & az    & FLEURS~\cite{conneau2022fleurs}                         & $10$         & $\mathbf{305}$     & $\mathbf{31}{\times}$ \\
\textit{Tamil}                      & ta    & Shrutilipi\textsuperscript{$\S$}                        & $790$        & $240$              & $0.30{\times}$ \\
Tagalog                             & tl    & FLEURS~\cite{conneau2022fleurs}                         & $10$         & $\mathbf{219}$     & $\mathbf{22}{\times}$ \\
Georgian                            & ka    & Common Voice 25~\cite{ardila2020common}                 & $168$        & $\mathbf{206}$     & $\mathbf{1.23}{\times}$ \\
\textit{Uyghur}                     & ug    & Common Voice 25~\cite{ardila2020common}                 & $451$        & $200$              & $0.44{\times}$ \\
Mongolian                           & mn    & Common Voice 25~\cite{ardila2020common}                 & $46$         & $\mathbf{181}$     & $\mathbf{3.93}{\times}$ \\
\textit{Kazakh}                     & kk    & KSD (OpenSLR 140)                                       & $554$        & $179$              & $0.32{\times}$ \\
Romansh                             & rm    & Common Voice 25~\cite{ardila2020common}                 & $8$          & $\mathbf{163}$     & $\mathbf{20}{\times}$ \\
\textit{Sinhala}                    & si    & OpenSLR 52                                              & $224$        & $154$              & $0.69{\times}$ \\
Hausa                               & ha    & BibleTTS~\cite{meyer2022bibletts}                       & $87$         & $\mathbf{126}$     & $\mathbf{1.45}{\times}$ \\
\textit{Marathi}                    & mr    & Shrutilipi\textsuperscript{$\S$}                 & $1{,}020$    & $114$              & $0.11{\times}$ \\
\textit{Urdu}                       & ur    & Shrutilipi\textsuperscript{$\S$}                 & $190$        & $86$               & $0.45{\times}$ \\
\textit{Telugu}                     & te    & Shrutilipi\textsuperscript{$\S$}                 & $390$        & $77$               & $0.20{\times}$ \\
\textit{Bengali}                    & bn    & Shrutilipi\textsuperscript{$\S$}                        & $440$        & $73$               & $0.17{\times}$ \\
\textit{Swedish}                    & sv    & RixVox                                                  & $5{,}493$    & $66$               & $0.01{\times}$ \\
\textit{Nepali}                     & ne    & OpenSLR 54                                              & $165$        & $64$               & $0.39{\times}$ \\
Irish                               & ga    & ABAIR-\'{E}IST                                          & $46$         & $\mathbf{61}$      & $\mathbf{1.32}{\times}$ \\
\textit{Odia}                       & or    & Shrutilipi\textsuperscript{$\S$}                 & $600$        & $58$               & $0.10{\times}$ \\
\textit{Malayalam}                  & ml    & Shrutilipi\textsuperscript{$\S$}                 & $360$        & $57$               & $0.16{\times}$ \\
Samoan                              & sm    & {--}                                                    & {--}         & $\mathbf{56}$      & \textbf{new} \\
Assamese                            & as    & Common Voice 25~\cite{ardila2020common}                 & $3$          & $\mathbf{55}$      & $\mathbf{18}{\times}$ \\
\textit{Setswana}                   & tn    & NCHLT~\cite{barnard2014nchlt}                           & $56$         & $51$               & $0.90{\times}$ \\
Montenegrin                         & cnr   & {--}                                                    & {--}         & $\mathbf{48}$      & \textbf{new} \\
Mauritian Creole                    & mfe   & {--}                                                    & {--}         & $\mathbf{44}$      & \textbf{new} \\
Hebrew                              & he    & FLEURS~\cite{conneau2022fleurs}                         & $10$         & $\mathbf{42}$      & $\mathbf{4.18}{\times}$ \\
Igbo                                & ig    & FLEURS~\cite{conneau2022fleurs}                         & $12$         & $\mathbf{41}$      & $\mathbf{3.39}{\times}$ \\
Amharic                             & am    & ALFFA (OpenSLR 25)                                      & $22$         & $\mathbf{40}$      & $\mathbf{1.80}{\times}$ \\
Latin                               & la    & {--}                                                    & {--}         & $\mathbf{35}$      & \textbf{new} \\
\textit{Central Kurdish}            & ckb   & Common Voice 25~\cite{ardila2020common}                 & $137$        & $35$               & $0.26{\times}$ \\
Dogri                               & dgo   & {--}                                                    & {--}         & $\mathbf{35}$      & \textbf{new} \\
Inuktitut                           & iu    & {--}                                                    & {--}         & $\mathbf{34}$      & \textbf{new} \\
\textit{Uzbek}                      & uz    & ISSAI USC                                               & $105$        & $34$               & $0.32{\times}$ \\
\textit{Kinyarwanda}                & rw    & Common Voice 25~\cite{ardila2020common}                 & $2{,}002$    & $32$               & $0.02{\times}$ \\
\textit{Kannada}                    & kn    & Shrutilipi\textsuperscript{$\S$}                 & $460$        & $30$               & $0.07{\times}$ \\
\textit{Persian}                    & fa    & Common Voice 25~\cite{ardila2020common}                 & $373$        & $28$               & $0.07{\times}$ \\
\textit{Gujarati}                   & gu    & Shrutilipi\textsuperscript{$\S$}                 & $460$        & $27$               & $0.06{\times}$ \\
\textit{Belarusian}                 & be    & Common Voice 25~\cite{ardila2020common}                 & $1{,}816$    & $24$               & $0.01{\times}$ \\
\textit{Dhivehi}                    & dv    & Common Voice 25~\cite{ardila2020common}                 & $38$         & $20$               & $0.53{\times}$ \\
\textit{Afrikaans}                  & af    & NCHLT~\cite{barnard2014nchlt}                           & $56$         & $20$               & $0.36{\times}$ \\
\textit{Zulu}                       & zu    & NCHLT~\cite{barnard2014nchlt}                           & $56$         & $19$               & $0.34{\times}$ \\
Shona                               & sn    & FLEURS~\cite{conneau2022fleurs}                         & $12$         & $\mathbf{18}$      & $\mathbf{1.52}{\times}$ \\
\textit{Oromo}                      & om    & Sagalee                                                 & $100$        & $16$               & $0.16{\times}$ \\
\textit{Esperanto}                  & eo    & Common Voice 25~\cite{ardila2020common}                 & $1{,}441$    & $15$               & $0.01{\times}$ \\
Tigrinya                            & ti    & {--}                                                    & {--}         & $\mathbf{14}$      & \textbf{new} \\
\textit{Xhosa}                      & xh    & NCHLT~\cite{barnard2014nchlt}                           & $56$         & $10$               & $0.18{\times}$ \\
\textit{Catalan}                    & ca    & Common Voice 25~\cite{ardila2020common}                 & $3{,}360$    & $1{,}171$         & $0.35{\times}$ \\
Spanish                             & es    & MLS~\cite{pratap2020mls}                                & $917$        & $\mathbf{6{,}792}$ & $\mathbf{7.41}{\times}$ \\
Dutch                               & nl    & MLS~\cite{pratap2020mls}                                & $1{,}554$    & $\mathbf{4{,}498}$ & $\mathbf{2.89}{\times}$ \\
Polish                              & pl    & ParlaSpeech 3.0~\cite{ljubesic2025parlaspeech3}         & $1{,}009$    & $\mathbf{2{,}732}$ & $\mathbf{2.71}{\times}$ \\
Czech                               & cs    & ParCzech4Speech                                         & $2{,}695$    & $\mathbf{3{,}717}$ & $\mathbf{1.38}{\times}$ \\
Cantonese                           & yue   & Common Voice 25~\cite{ardila2020common}                 & $211$        & $\mathbf{1{,}944}$ & $\mathbf{9.21}{\times}$ \\
Luxembourgish                       & lb    & RTL.lu ASR\textsuperscript{$\dagger$}                   & $67$         & $\mathbf{1{,}805}$ & $\mathbf{27}{\times}$ \\
Portuguese                          & pt    & CORAA                                                   & $291$        & $\mathbf{1{,}764}$ & $\mathbf{6.06}{\times}$ \\
Romanian                            & ro    & VoxPopuli~\cite{wang2021voxpopuli}                      & $89$         & $\mathbf{1{,}746}$ & $\mathbf{20}{\times}$ \\
\textit{English}                    & en    & MLS~\cite{pratap2020mls}                                & $44{,}659$   & $5{,}312$          & $0.12{\times}$ \\
Hindi                               & hi    & Shrutilipi~\cite{bhogale2022shrutilipi}                 & $1{,}620$    & $\mathbf{1{,}707}$ & $\mathbf{1.05}{\times}$ \\
Kreol Seselwa                       & crs   & {--}                                                    & {--}         & $\mathbf{1{,}602}$ & \textbf{new} \\
Russian                             & ru    & Common Voice 25~\cite{ardila2020common}                 & $252$        & $\mathbf{1{,}537}$ & $\mathbf{6.10}{\times}$ \\
Mandarin                            & zh    & Common Voice 25~\cite{ardila2020common}                 & $427$        & $\mathbf{1{,}482}$ & $\mathbf{3.47}{\times}$ \\
Khmer                               & km    & FLEURS~\cite{conneau2022fleurs}                         & $10$         & $\mathbf{1{,}323}$ & $\mathbf{132}{\times}$ \\
Japanese                            & ja    & Common Voice 25~\cite{ardila2020common}                 & $372$        & $\mathbf{1{,}387}$ & $\mathbf{3.73}{\times}$ \\
Korean                              & ko    & Zeroth (OpenSLR 40)\textsuperscript{$\ddagger$}         & $53$         & $\mathbf{1{,}454}$ & $\mathbf{27}{\times}$ \\
Armenian                            & hy    & OpenSLR 160                                             & $70$         & $\mathbf{1{,}139}$ & $\mathbf{16}{\times}$ \\
Turkish                             & tr    & Common Voice 25~\cite{ardila2020common}                 & $129$        & $\mathbf{1{,}008}$ & $\mathbf{7.81}{\times}$ \\
Arabic                              & ar    & Common Voice 25~\cite{ardila2020common}                 & $92$         & $\mathbf{1{,}001}$ & $\mathbf{11}{\times}$ \\
Swahili                             & sw    & Common Voice 25~\cite{ardila2020common}                 & $392$        & $\mathbf{1{,}006}$ & $\mathbf{2.57}{\times}$ \\
Thai                                & th    & Common Voice 25~\cite{ardila2020common}                 & $173$        & $\mathbf{1{,}176}$ & $\mathbf{6.80}{\times}$ \\
Hungarian                           & hu    & Common Voice 25~\cite{ardila2020common}                 & $133$        & $\mathbf{1{,}350}$ & $\mathbf{10}{\times}$ \\
French                              & fr    & EuroSpeech~\cite{pfisterer2025eurospeech}               & $2{,}250$    & $\mathbf{6{,}029}$ & $\mathbf{2.68}{\times}$ \\
\textit{German}                     & de    & EuroSpeech~\cite{pfisterer2025eurospeech}               & $2{,}184$    & $1{,}907$          & $0.87{\times}$ \\
\end{longtable}
\end{small}

\noindent\textsuperscript{$\dagger$}RTL.lu ASR dataset (CC BY-NC-ND 4.0): \url{https://huggingface.co/datasets/Lemswasabi/luxembourgish-asr-rtl-lu}. \\
\noindent\textsuperscript{$\ddagger$}KsponSpeech (969\,h) is excluded: AIHub restricts access to Korean-resident applicants only and prohibits redistribution between institutions~\cite{bang2020ksponspeech}. Zeroth Korean (OpenSLR 40, CC BY 4.0) is the largest freely redistributable prior. \\
\noindent\textsuperscript{$\S$}Shrutilipi~\cite{bhogale2022shrutilipi} covers Indian Bengali and Indian Tamil (AIR news broadcasts, CC BY 4.0). WorldSpeech's \texttt{bd\_bn} is Bangladeshi Bengali via VOA and \texttt{ta\_lk} is Sri Lankan Tamil, distinct dialects and domains.

\section{Per-source license census}
\label{app:licenses}

Table~\ref{tab:licenses} lists the legal basis for redistribution of each \textsc{WorldSpeech} configuration. License names are hyperlinked to the cited primary source. Sorted by aligned hours descending; the six RFA and RFE/RL configurations appear at the bottom as their redistribution status is pending confirmation.

\begin{small}
\setlength{\tabcolsep}{3pt}
\renewcommand{\arraystretch}{0.92}
\begin{longtable}{@{}l p{2.4cm} p{1.9cm} p{3.6cm} p{4.2cm}@{}}
\caption{Per-configuration license census for \textsc{WorldSpeech}. Config codes match \texttt{disco-eth/WorldSpeech} on HuggingFace. License names link to the cited legal document.}
\label{tab:licenses} \\
\toprule
Config & Country & Language & Source & Legal basis \\
\midrule
\endfirsthead
\multicolumn{5}{l}{\emph{(Table~\ref{tab:licenses} continued)}} \\
\toprule
Config & Country & Language & Source & Legal basis \\
\midrule
\endhead
\midrule
\multicolumn{5}{r}{\emph{(continued on next page)}} \\
\endfoot
\bottomrule
\endlastfoot
  \texttt{pl\_pl} & Poland & Polish & Sejm & \href{https://isap.sejm.gov.pl/isap.nsf/DocDetails.xsp?id=WDU19940240083}{Polish Copyright Act Art.~4(2)} \\
  \texttt{cs\_cz} & Czech Republic & Czech & Chamber of Deputies (PSP) & \href{https://www.zakonyprolidi.cz/cs/2000-121}{Czech Copyright Act S.~3(1)(c)} \\
  \texttt{nl\_nl} & Netherlands & Dutch & Tweede Kamer & \href{https://wetten.overheid.nl/BWBR0001886}{Dutch Author's Rights Act Art.~11} \\
  \texttt{cz\_cs} & Czech Republic & Czech & Chamber of Deputies (PSP) & \href{https://www.zakonyprolidi.cz/cs/2000-121}{Czech Copyright Act S.~3(1)(c)} \\
  \texttt{es\_es} & Spain & Spanish & Congreso de los Diputados & \href{https://www.boe.es/buscar/act.php?id=BOE-A-1996-8930}{Spanish Copyright Law Art.~13} \\
  \texttt{yue\_hk} & Hong Kong & Cantonese & Legislative Council & \href{https://www.elegislation.gov.hk/hk/cap528}{HK Copyright Ordinance Cap.~528 (Speaker's permission convention)} \\
  \texttt{lb\_lu} & Luxembourg & Luxembourgish & Chamber of Deputies & \href{https://legilux.public.lu/eli/etat/leg/loi/2001/04/18/n1/jo}{Luxembourg Copyright Law 2001} \\
  \texttt{pt\_br} & Brazil & Portuguese & Federal Senate & \href{https://www.planalto.gov.br/ccivil_03/leis/l9610.htm}{Brazilian Copyright Law Art.~8(IV)} \\
  \texttt{es\_cl} & Chile & Spanish & Chamber of Deputies + Senate & \href{https://www.bcn.cl/leychile/navegar?idNorma=28979}{Chilean Copyright Law Art.~71-S} \\
  \texttt{hi\_in} & India & Hindi & Vidhan Sabhas + Mann Ki Baat & \href{https://copyright.gov.in/documents/copyrightact1957.pdf}{Indian Copyright Act 1957 S.~52(1)(q)} \\
  \texttt{crs\_sc} & Seychelles & Kreol Seselwa & National Assembly & \href{https://seylii.org/akn/sc/act/2014/5/eng@2015-12-31}{Seychelles Copyright Act 2014 (No.~5/2014)} \\
  \texttt{ru\_ru} & Russia & Russian & State Duma & \href{https://www.consultant.ru/document/cons_doc_LAW_64629/}{Russian Civil Code Art.~1259(6)} \\
  \texttt{zh\_tw} & Taiwan & Mandarin & Legislative Yuan IVOD & \href{https://law.moj.gov.tw/ENG/LawClass/LawAll.aspx?pcode=J0070017}{Taiwan Copyright Act Art.~9(1)} \\
  \texttt{ja\_jp} & Japan & Japanese & LibriVox + Aozora Bunko & \href{https://creativecommons.org/publicdomain/zero/1.0/}{CC0 / Public Domain Dedication} \\
  \texttt{ko\_kr} & South Korea & Korean & National Assembly & \href{https://elaw.klri.re.kr/kor_service/lawView.do?hseq=61780}{Korean Copyright Act Art.~7(1)} \\
  \texttt{hy\_am} & Armenia & Armenian & National Assembly & \href{https://www.wipo.int/wipolex/en/legislation/details/15105}{Armenian Copyright Law Art.~4(1)(c)} \\
  \texttt{fr\_ca} & Canada (Quebec) & French & Quebec National Assembly & \href{https://www.assnat.qc.ca/en/abc-assnat/footer/avis-juridiques.html}{Quebec NA Speaker's permission / parliamentary privilege} \\
  \texttt{de\_at} & Austria & German & National Council + Federal Council & \href{https://www.ris.bka.gv.at/GeltendeFassung.wxe?Abfrage=Bundesnormen&Gesetzesnummer=10001848}{Austrian UrhG S.~7} \\
  \texttt{ro\_md} & Moldova & Romanian & Parliament of Moldova & \href{https://www.legis.md/cautare/getResults?doc_id=131988}{Moldovan Copyright Law Art.~8(f)} \\
  \texttt{tr\_tr} & Turkey & Turkish & Grand National Assembly & \href{https://www.mevzuat.gov.tr/MevzuatMetin/1.3.5846.pdf}{Turkish Copyright Law (FSEK 5846) Art.~31} \\
  \texttt{nl\_be} & Belgium & Dutch & Flemish Parliament & \href{https://www.ejustice.just.fgov.be/eli/loi/2013/02/28/2013A11134/justel}{Belgian Code of Economic Law Art.~XI.172} \\
  \texttt{es\_mx} & Mexico & Spanish & Mexico City Congress + SCJN & \href{https://www.diputados.gob.mx/LeyesBiblio/pdf/LFDA.pdf}{Mexican Copyright Law Art.~14(VIII)} \\
  \texttt{es\_uy} & Uruguay & Spanish & Chamber of Representatives + Senate & \href{https://www.impo.com.uy/bases/leyes/9739-1937}{Uruguayan Copyright Law Art.~45 numeral~5} \\
  \texttt{sw\_tz} & Tanzania & Swahili & Bunge of Tanzania & \href{https://www.tanzlii.org/tz/legislation/act/1999/7}{Tanzania Copyright Act Cap.~218 S.~7} \\
  \texttt{ro\_ro} & Romania & Romanian & Senate of Romania & \href{https://legislatie.just.ro/Public/DetaliiDocumentAfis/13816}{Romanian Copyright Law No.~8/1996} \\
  \texttt{th\_th} & Thailand & Thai & Parliament of Thailand & \href{https://www.ipthailand.go.th/images/004/Copyright_Act_2537.pdf}{Thai Copyright Act B.E.~2537 S.~7(2)} \\
  \texttt{hu\_hu} & Hungary & Hungarian & National Assembly & \href{https://njt.hu/jogszabaly/1999-76-00-00}{Hungarian Copyright Act S.~1(4)} \\
  \texttt{en\_au} & Australia & English & Australian Parliament House & \href{https://www.aph.gov.au/Help/Disclaimer_Privacy_Copyright}{CC BY-NC-ND 4.0} \\
  \texttt{en\_nz} & New Zealand & English & House of Representatives & \href{https://www.legislation.govt.nz/act/public/1994/0143/latest/DLM345939.html}{NZ Copyright Act 1994 S.~27} \\
  \texttt{ms\_my} & Malaysia & Malay & Parliament of Malaysia & \href{https://www.myipo.gov.my/wp-content/uploads/2025/09/Copyright-Act-1987-Act-332.pdf}{Malaysian Copyright Act 1987 S.~3} \\
  \texttt{el\_cy} & Cyprus & Greek & House of Representatives & \href{https://www.cylaw.org/nomoi/enop/non-ind/2006_1_59/full.html}{Cyprus Copyright Law S.~7(2)} \\
  \texttt{es\_pe} & Peru & Spanish & Congress of the Republic & \href{https://www.indecopi.gob.pe/documents/20795/203018/DL822.pdf}{Peruvian Copyright Law Art.~15} \\
  \texttt{az\_voa} & Azerbaijan & Azerbaijani & Voice of America & \href{https://www.copyright.gov/title17/92chap1.html\#105}{17 U.S.C. S.~105} \\
  \texttt{iq\_ar} & Iraq & Arabic & Council of Representatives & \href{https://www.wipo.int/wipolex/en/legislation/details/10345}{Iraqi Copyright Law No.~3 of 1971 Art.~6} \\
  \texttt{en\_zm} & Zambia & English & National Assembly of Zambia & \href{https://www.wipo.int/wipolex/en/legislation/details/21204}{Zambia Copyright and Performance Rights Act Cap.~406 S.~8(2)} \\
  \texttt{am\_hy} & Armenia & Armenian & National Assembly & \href{https://www.wipo.int/wipolex/en/legislation/details/15105}{Armenian Copyright Law Art.~4(1)(c)} \\
  \texttt{sq\_al} & Albania & Albanian & Assembly of Albania & \href{https://qbz.gov.al/eli/fz/2016/131/ac97f98c-2f9b-4b21-8d98-ecc9de7b5310}{Albanian Copyright Law Art.~8} \\
  \texttt{es\_ar} & Argentina & Spanish & Chamber of Deputies + Senate & \href{https://servicios.infoleg.gob.ar/infolegInternet/anexos/40000-44999/42755/texact.htm}{Argentine Copyright Law Art.~12} \\
  \texttt{es\_pr} & Puerto Rico & Spanish & House of Representatives + Senate & \href{https://www.copyright.gov/title17/92chap1.html\#105}{US Government works (official acts)} \\
  \texttt{bh\_ar} & Bahrain & Arabic & Council of Representatives & \href{https://www.wipo.int/wipolex/en/legislation/details/19867}{Bahraini Copyright Law No.~22 of 2006 Art.~4} \\
  \texttt{tl\_ph} & Philippines & Tagalog & House + Senate & \href{https://ipophl.gov.ph/laws-issuances/republic-acts/republic-act-no-8293/}{Philippine IP Code S.~176.1} \\
  \texttt{ka\_ge} & Georgia & Georgian & Parliament of Georgia & \href{https://matsne.gov.ge/en/document/view/1755592}{Georgian Copyright Law Art.~8} \\
  \texttt{ta\_lk} & Sri Lanka & Tamil & Parliament of Sri Lanka & \href{https://www.nipo.gov.lk/web/images/pdf_downloads/Intellectual_Property_Act_No_36_of_2003.pdf}{Sri Lanka IP Act No.~36/2003 S.~8} \\
  \texttt{mn\_mn} & Mongolia & Mongolian & State Great Khural & \href{https://www.wipo.int/wipolex/en/legislation/details/22459}{Mongolian Copyright Law (2021) Art.~8} \\
  \texttt{kk\_kz} & Kazakhstan & Kazakh & Mazhilis & \href{https://adilet.zan.kz/rus/docs/Z960000006_}{Kazakh Copyright Law Art.~8} \\
  \texttt{sq\_xk} & Kosovo & Albanian & Assembly of Kosovo & \href{https://mkrs-ks.org/repository/docs/LAW_NO._08_L-205_ON_COPYRIGHT_AND_RELATED_RIGHTS.pdf}{Kosovo Copyright Law 08/L-205 (2023)} \\
  \texttt{en\_ke} & Kenya & English & Parliament of Kenya & \href{https://copyright.go.ke/sites/default/files/downloads/CopyrightAct12of2001.pdf}{Kenya Copyright Act Cap.~130 S.~26(1)(d)} \\
  \texttt{rm\_ch} & Switzerland & Romansh & RTR (Radio Televisiun Svizra Rumantscha) & \href{https://www.rtr.ch}{Research data-sharing agreement} \\
  \texttt{es\_co} & Colombia & Spanish & House + Senate & \href{https://www.secretariasenado.gov.co/senado/basedoc/ley_0023_1982.html}{Colombian Copyright Law (Ley 23/1982) Art.~41} \\
  \texttt{si\_lk} & Sri Lanka & Sinhala & Parliament of Sri Lanka & \href{https://www.nipo.gov.lk/web/images/pdf_downloads/Intellectual_Property_Act_No_36_of_2003.pdf}{Sri Lanka IP Act No.~36/2003 S.~8} \\
  \texttt{en\_za} & South Africa & English & Parliament of South Africa & \href{https://www.saflii.org/za/legis/consol_act/ca1978120/}{SA Copyright Act 1978 S.~12(8)} \\
  \texttt{es\_py} & Paraguay & Spanish & Chamber of Deputies + Senate & \href{https://www.ip.gov.py/ip/marco-legal/propiedad-intelectual/}{Paraguayan Copyright Law Art.~4} \\
  \texttt{en\_sl} & Sierra Leone & English & Parliament of Sierra Leone & \href{https://www.wipo.int/wipolex/en/legislation/details/14529}{Sierra Leone Copyright Act 2011 (No.~8 of 2011)} \\
  \texttt{id\_id} & Indonesia & Indonesian & Voice of America & \href{https://www.copyright.gov/title17/92chap1.html\#105}{17 U.S.C. S.~105} \\
  \texttt{dz\_ar} & Algeria & Arabic & National People's Assembly & \href{https://www.wipo.int/wipolex/en/legislation/details/1194}{Algerian Copyright Ord.~03-05 Art.~11} \\
  \texttt{ma\_ar} & Morocco & Arabic & House of Representatives & \href{https://www.wipo.int/wipolex/en/legislation/details/19766}{Moroccan Copyright Law No.~2-00 Art.~8} \\
  \texttt{ha\_td} & Chad & Hausa & Voice of America & \href{https://www.copyright.gov/title17/92chap1.html\#105}{17 U.S.C. S.~105} \\
  \texttt{sv\_ax} & Aland Islands & Swedish & Lagting & \href{https://www.finlex.fi/en/laki/kaannokset/1961/en19610404}{Finnish Copyright Act Ch.~1 S.~9} \\
  \texttt{ga\_ie} & Ireland & Irish & Houses of the Oireachtas & \href{https://www.oireachtas.ie/en/copyright-and-reuse/}{Oireachtas (Open Data) PSI Licence (CC BY 4.0)} \\
  \texttt{ne\_np} & Nepal & Nepali & Federal Parliament & \href{https://www.nepalcopyright.gov.np/downloadfile/The\%20Copyright\%20Act_1315912777_1573122046.pdf}{Nepal Copyright Act 2059 S.~4} \\
  \texttt{ws\_sm} & Samoa & Samoan & Legislative Assembly & \href{https://www.wipo.int/wipolex/en/legislation/details/13393}{Samoa Copyright Act 1998 S.~5(b)} \\
  \texttt{ha\_ng} & Nigeria & Hausa & Voice of America & \href{https://www.copyright.gov/title17/92chap1.html\#105}{17 U.S.C. S.~105} \\
  \texttt{bw\_tn} & Botswana & Tswana & National Assembly & \href{https://www.wipo.int/wipolex/en/legislation/details/9583}{Botswana Copyright and Neighbouring Rights Act S.~6(2)(b)} \\
  \texttt{tn\_bw} & Botswana & Tswana & National Assembly & \href{https://www.wipo.int/wipolex/en/legislation/details/9583}{Botswana Copyright and Neighbouring Rights Act S.~6(2)(b)} \\
  \texttt{cnr\_me} & Montenegro & Montenegrin & Parliament of Montenegro & \href{https://www.wipo.int/wipolex/en/legislation/details/22248}{Montenegrin Copyright Law Art.~9} \\
  \texttt{bn\_bd} & Bangladesh & Bengali & Jatiya Sangsad & \href{https://www.wipo.int/wipolex/en/legislation/details/11172}{Bangladesh Copyright Act 2000 S.~72} \\
  \texttt{mfe\_mu} & Mauritius & Morisyen & National Assembly & \href{https://www.wipo.int/wipolex/en/legislation/details/15447}{Mauritius Copyright Act 2014 S.~5} \\
  \texttt{il\_he} & Israel & Hebrew & LibriVox + Ben-Yehuda Project & \href{https://creativecommons.org/publicdomain/zero/1.0/}{CC0 / Public Domain Dedication} \\
  \texttt{ig\_ng} & Nigeria & Igbo & Voice of America & \href{https://www.copyright.gov/title17/92chap1.html\#105}{17 U.S.C. S.~105} \\
  \texttt{et\_am\_voa} & Ethiopia & Amharic & Voice of America & \href{https://www.copyright.gov/title17/92chap1.html\#105}{17 U.S.C. S.~105} \\
  \texttt{el\_gr} & Greece & Greek & Hellenic Parliament & \href{https://www.wipo.int/wipolex/en/legislation/details/15581}{Greek Copyright Law 2121/1993 Art.~2(5)} \\
  \texttt{la\_va} & Vatican & Latin & LibriVox & \href{https://creativecommons.org/publicdomain/zero/1.0/}{CC0 / Public Domain Dedication} \\
  \texttt{ckb\_iq} & Iraqi Kurdistan & Central Kurdish & Kurdistan Parliament & \href{https://www.wipo.int/wipolex/en/legislation/details/10345}{Iraqi Copyright Law No.~3 of 1971 Art.~6} \\
  \texttt{ca\_iu} & Canada (Nunavut) & Inuktitut & Legislative Assembly of Nunavut & \href{https://www.assembly.nu.ca/important-notices}{Nunavut Leg.~Assembly terms of use / parliamentary privilege} \\
  \texttt{cd\_fr} & DR Congo & French & ICC trials (Lubanga et al.) & \href{https://www.icc-cpi.int/court-record}{Public court records (ICC)} \\
  \texttt{ir\_fa} & Iran & Persian & Voice of America & \href{https://www.copyright.gov/title17/92chap1.html\#105}{17 U.S.C. S.~105} \\
  \texttt{by\_be} & Belarus & Belarusian & Knihi.com archive & \href{https://www.wipo.int/wipolex/en/legislation/details/22302}{Belarus Copyright Law 262-Z (2011) Art.~7 (author's rights expired)} \\
  \texttt{eg\_ar} & Egypt & Arabic & House of Representatives & \href{https://www.wipo.int/wipolex/en/legislation/details/22066}{Egyptian IP Law No.~82 of 2002 Art.~141} \\
  \texttt{mv\_dv} & Maldives & Dhivehi & People's Majlis & \href{https://www.wipo.int/wipolex/en/legislation/details/9517}{Maldives Copyright Act (Law No.~23/2010) S.~6(b)} \\
  \texttt{za\_af} & South Africa & Afrikaans & Parliament of South Africa & \href{https://www.saflii.org/za/legis/consol_act/ca1978120/}{SA Copyright Act 1978 S.~12(8)} \\
  \texttt{za\_zu} & South Africa & Zulu & Parliament of South Africa & \href{https://www.saflii.org/za/legis/consol_act/ca1978120/}{SA Copyright Act 1978 S.~12(8)} \\
  \texttt{rw\_rw} & Rwanda & Kinyarwanda & Chamber of Deputies & \href{https://www.wipo.int/wipolex/en/legislation/details/5249}{Rwandan IP Law No.~31/2009 Art.~198} \\
  \texttt{et\_om} & Ethiopia & Oromo & Voice of America & \href{https://www.copyright.gov/title17/92chap1.html\#105}{17 U.S.C. S.~105} \\
  \texttt{rw\_voa} & Rwanda & Kinyarwanda & Voice of America & \href{https://www.copyright.gov/title17/92chap1.html\#105}{17 U.S.C. S.~105} \\
  \texttt{xx\_eo} & International & Esperanto & LibriVox & \href{https://creativecommons.org/publicdomain/zero/1.0/}{CC0 / Public Domain Dedication} \\
  \texttt{ti\_voa} & Eritrea & Tigrinya & Voice of America & \href{https://www.copyright.gov/title17/92chap1.html\#105}{17 U.S.C. S.~105} \\
  \texttt{ci\_fr} & Cote d'Ivoire & French & ICC trial (Gbagbo et al.) & \href{https://www.icc-cpi.int/court-record}{Public court records (ICC)} \\
  \texttt{un\_ar} & United Nations & Arabic & UN General Assembly + SC & \href{https://www.un.org/en/about-us/terms-of-use}{UN parliamentary documents (ODS), non-commercial reproduction with credit} \\
  \texttt{en\_jm} & Jamaica & English & Parliament of Jamaica & \href{https://moj.gov.jm/sites/default/files/laws/Copyright\%20Act.pdf}{Jamaican Copyright Act S.~6(5)} \\
  \texttt{za\_xh} & South Africa & Xhosa & Parliament of South Africa & \href{https://www.saflii.org/za/legis/consol_act/ca1978120/}{SA Copyright Act 1978 S.~12(8)} \\
  \texttt{sa\_ar} & Saudi Arabia & Arabic & Public-domain Arabic audio (archive.org) & \href{https://archive.org/about/terms}{Internet Archive ToU (per-item license)} \\
  \texttt{by\_ru} & Belarus & Russian & Presidential addresses & \href{https://www.wipo.int/wipolex/en/legislation/details/22302}{Belarus Copyright Law 262-Z (2011) Art.~7(2), official documents not protected} \\
  \texttt{nz\_mi} & New Zealand & Maori & House of Representatives & \href{https://www.legislation.govt.nz/act/public/1994/0143/latest/DLM345939.html}{NZ Copyright Act 1994 S.~27} \\
  \texttt{gr\_grc} & Greece & Ancient Greek & LibriVox & \href{https://creativecommons.org/publicdomain/zero/1.0/}{CC0 / Public Domain Dedication} \\
  \texttt{tn\_za} & South Africa & Tswana & Parliament of South Africa & \href{https://www.saflii.org/za/legis/consol_act/ca1978120/}{SA Copyright Act 1978 S.~12(8)} \\
  \texttt{za\_nso} & South Africa & Northern Sotho & Parliament of South Africa & \href{https://www.saflii.org/za/legis/consol_act/ca1978120/}{SA Copyright Act 1978 S.~12(8)} \\
  \texttt{za\_st} & South Africa & Sesotho & Parliament of South Africa & \href{https://www.saflii.org/za/legis/consol_act/ca1978120/}{SA Copyright Act 1978 S.~12(8)} \\
  \texttt{za\_ts} & South Africa & Tsonga & Parliament of South Africa & \href{https://www.saflii.org/za/legis/consol_act/ca1978120/}{SA Copyright Act 1978 S.~12(8)} \\
  \texttt{za\_nr} & South Africa & S. Ndebele & Parliament of South Africa & \href{https://www.saflii.org/za/legis/consol_act/ca1978120/}{SA Copyright Act 1978 S.~12(8)} \\
  \texttt{za\_ss} & South Africa & Swati & Parliament of South Africa & \href{https://www.saflii.org/za/legis/consol_act/ca1978120/}{SA Copyright Act 1978 S.~12(8)} \\
  \texttt{za\_ve} & South Africa & Venda & Parliament of South Africa & \href{https://www.saflii.org/za/legis/consol_act/ca1978120/}{SA Copyright Act 1978 S.~12(8)} \\
  \texttt{kh\_km} & Cambodia & Khmer & Radio Free Asia & \href{https://www.rfa.org/about/info/terms-of-use.html}{RFA ToU (non-commercial reuse with attribution)} \\
  \texttt{lo\_la} & Laos & Lao & Radio Free Asia & \href{https://www.rfa.org/about/info/terms-of-use.html}{RFA ToU (non-commercial reuse with attribution)} \\
  \texttt{vn\_vi} & Vietnam & Vietnamese & Radio Free Asia & \href{https://www.rfa.org/about/info/terms-of-use.html}{RFA ToU (non-commercial reuse with attribution)} \\
  \texttt{hk\_yue\_rfa} & Hong Kong & Cantonese (RFA) & Radio Free Asia & \href{https://www.rfa.org/about/info/terms-of-use.html}{RFA ToU (non-commercial reuse with attribution)} \\
  \texttt{cn\_ug} & China & Uyghur & Radio Free Asia & \href{https://www.rfa.org/about/info/terms-of-use.html}{RFA ToU (non-commercial reuse with attribution)} \\
  \texttt{uz\_uz} & Uzbekistan & Uzbek & Ozodlik (RFE/RL) & \href{https://about.rferl.org/use-our-content/}{RFE/RL ToU (note: AI-training restrictions; permission pending)} \\
\end{longtable}
\end{small}

\noindent The six configurations sourced from Radio Free Asia and Radio Free Europe / Radio Liberty are released as alignment metadata only (transcripts, source URLs, and segment timestamps), with audio retrievable by the user from the original source. This split-release model follows established practice for speech corpora that include third-party audio whose redistribution terms are not fully permissive~\cite{kahn2020librilight}.

\section{ASR Fine-tuning on \textsc{WorldSpeech}}
\label{app:fine_tuning}
\label{app:asr_results}

\begin{table}[!htbp]
\centering
\small
\setlength{\tabcolsep}{4pt}
\renewcommand{\arraystretch}{1.05}
\caption{ASR fine-tuning results with \texttt{whisper-large-v3-turbo} backbone. Rows marked $^{\ast}$ use the \textsc{WorldSpeech} held-out test split (test benchmark unavailable), all others use FLEURS test.}
\label{tab:asr_results_v3}
\begin{tabularx}{0.99\textwidth}{l l Y Y Y Y Y Y}
\toprule
Language & Benchmark & WER\textsubscript{base} & WER\textsubscript{FT} & WER $\Delta$ & CER\textsubscript{base} & CER\textsubscript{FT} & CER $\Delta$ \\
\midrule
Samoan$^{\ast}$    & \textsc{WorldSpeech} & $4.719$ & $0.393$ & $-91.7\%$ & $3.870$ & $0.258$ & $-93.3\%$ \\
Lao                & FLEURS & $2.469$ & $0.756$ & $-69.4\%$ & $2.407$ & $0.275$ & $-88.6\%$ \\
Kreol Seselwa$^{\ast}$ & \textsc{WorldSpeech} & $1.633$ & $0.704$ & $-56.9\%$ & $1.182$ & $0.512$ & $-56.7\%$ \\
Romansh$^{\ast}$   & \textsc{WorldSpeech} & $1.314$ & $0.165$ & $-87.5\%$ & $0.822$ & $0.049$ & $-94.1\%$ \\
Georgian           & FLEURS & $1.070$ & $0.480$ & $-55.1\%$ & $1.090$ & $0.200$ & $-81.7\%$ \\
Burmese            & FLEURS & $1.006$ & $0.390$ & $-61.2\%$ & $1.266$ & $0.282$ & $-77.8\%$ \\
Luxembourgish      & FLEURS & $0.946$ & $0.284$ & $-70.0\%$ & $0.390$ & $0.091$ & $-76.7\%$ \\
Arabic (Bahrain)$^{\ast}$ & \textsc{WorldSpeech} & $0.617$ & $0.302$ & $-51.1\%$ & $0.268$ & $0.210$ & $-21.6\%$ \\
Albanian$^{\ast}$  & \textsc{WorldSpeech} & $0.554$ & $0.236$ & $-57.4\%$ & $0.243$ & $0.143$ & $-41.1\%$ \\
Armenian           & FLEURS & $0.427$ & $0.178$ & $-58.3\%$ & $0.092$ & $0.090$ & $-2.6\%$  \\
Swahili            & FLEURS & $0.328$ & $0.196$ & $-40.2\%$ & $0.090$ & $0.075$ & $-16.7\%$ \\
\bottomrule
\end{tabularx}
\end{table}

\section{Codebase}
\label{app:codebase}

The full \textsc{WorldSpeech} codebase is released at \url{https://github.com/ETH-DISCO/worldspeech}. The repository contains the per-source data sourcing scripts (one collection per country-language configuration), the alignment pipeline including the iterative refinement loop of Section~\ref{sec:iterative}, and the ASR fine-tuning and evaluation scripts used in Section~\ref{sec:experiments}. Together these reproduce the data construction and the fine-tuning results reported in Table~\ref{tab:asr_results_v3} and the ablation of Figure~\ref{fig:hours_ablation}.

\section{Compute resources}
\label{app:compute}

Total compute consumed by the project was approximately $19{,}400$ GPU-hours and $33{,}000$ CPU-hours, distributed across $21{,}183$ GPU jobs and $34{,}139$ CPU jobs. Workload is dominated by per-segment ASR transcription during alignment, with smaller contributions from pilot ASR runs, the per-language fine-tuning of Section~\ref{sec:experiments}, the hours-vs-WER ablation of Figure~\ref{fig:hours_ablation}, and the iterative alignment refinement of Section~\ref{sec:iterative}. Jobs ran on a cluster of NVIDIA RTX (Ampere and Ada generations) and A6000/A100 GPUs.

\section{Iterative Alignment Refinement: Beyond the second pass}
\label{app:pass3}

Table~\ref{tab:pass3} reports the pass 3 yield of the iterative alignment refinement procedure (Section~\ref{sec:iterative}) for the nine languages of Figure~\ref{fig:iar_dumbbell}. After pass 2 the fine-tuned ASR has already absorbed most of the recoverable signal, so pass 3 fine-tunes a new ASR on the pass 1 plus pass 2 yield and re-aligns the residual unaligned audio. pass 3 returns much smaller additional hours, ranging from $+0.2\%$ (Flemish) to $+8.8\%$ (Burmese) with an average of $+4.3\%$, compared to a pass 1 to pass 2 average of $+95.4\%$. Languages whose initial ASR was already strong (Flemish, Armenian, Tamil) saturate after one refinement pass, while the languages with the weakest initial model (Burmese, Lao, Khmer) retain modest residual headroom but still gain less than a tenth of what pass 2 contributed. We therefore stop the refinement loop at pass 2 in the released corpus.

\begin{table}[!htbp]
\centering
\small
\setlength{\tabcolsep}{6pt}
\renewcommand{\arraystretch}{1.05}
\caption{Pass 3 yield for the nine iterative-alignment-refinement languages of Figure~\ref{fig:iar_dumbbell}. pass 1 is the initial model alignment, pass 2 the re-alignment with the pass-1 fine-tuned ASR, and pass 3 the re-alignment with a pass-2 fine-tuned ASR. The average pass 3 gain over pass 2 is $+4.3\%$, against $+95.4\%$ for pass 2 over pass 1.}
\label{tab:pass3}
\begin{tabular}{lrrrrrr}
\toprule
Language & Pass 1 & Pass 2 & P1$\to$P2 & Pass 3 & P2$\to$P3 & abs.\ added \\
\midrule
Burmese          & $287.3$ & $865.0$     & $+201.1\%$ & $941$     & $+8.8\%$ & $+76.0$ \\
Lao              & $296.4$ & $827.0$     & $+179.0\%$ & $893$     & $+8.0\%$ & $+66.0$ \\
Khmer            & $528.7$ & $1{,}323.0$ & $+150.2\%$ & $1{,}408$ & $+6.4\%$ & $+85.0$ \\
Kreol Seselwa    & $802.7$ & $1{,}602.3$ & $+99.6\%$  & $1{,}684$ & $+5.1\%$ & $+81.7$ \\
Sinhala          & $67.4$  & $154.0$     & $+128.5\%$ & $161$     & $+4.5\%$ & $+7.0$ \\
Bahraini Arabic  & $143.6$ & $272.5$     & $+89.8\%$  & $282$     & $+3.5\%$ & $+9.5$ \\
Tamil            & $134.3$ & $204.0$     & $+51.9\%$  & $207$     & $+1.5\%$ & $+3.0$ \\
Armenian         & $815.2$ & $1{,}138.9$ & $+39.7\%$  & $1{,}146$ & $+0.6\%$ & $+7.1$ \\
Flemish          & $803.6$ & $960.5$     & $+19.5\%$  & $962$     & $+0.2\%$ & $+1.5$ \\
\bottomrule
\end{tabular}
\end{table}

\end{document}